\PassOptionsToPackage{table}{xcolor}
\documentclass[11pt]{article}

\usepackage[final]{acl}

\usepackage{times}
\usepackage{latexsym}

\usepackage[T1]{fontenc}

\usepackage[utf8]{inputenc}

\usepackage{microtype}

\usepackage{inconsolata}

\usepackage{graphicx}

\usepackage{amsmath,amssymb,bbm}
\definecolor{myredd}{RGB}{176,35,24}
\definecolor{mygreenn}{RGB}{78,173,91}
\definecolor{mybluee}{RGB}{46,112,186}
\definecolor{myred}{RGB}{255,225,220}
\definecolor{mygreen}{RGB}{194,244,210} 
\definecolor{myblue}{RGB}{216,236,255}
\definecolor{byzantine}{rgb}{0.74, 0.2, 0.64}
\definecolor{table-blue}{RGB}{173, 216, 230}
\definecolor{mygrey}{HTML}{E8E8E8}
\definecolor{myblue}{HTML}{C8D9ED}
\definecolor{myorange}{HTML}{F2D7C3}
\definecolor{mypurple}{HTML}{BC689F}

\usepackage{booktabs}
\usepackage{multirow}
\usepackage{graphicx}
\usepackage{tikz}
\usetikzlibrary{shapes, backgrounds, positioning}
\usepackage{pgfplots}
\usepgfplotslibrary{groupplots}
\usetikzlibrary{decorations.pathreplacing, calc, decorations.pathmorphing, arrows.meta, shapes.geometric, arrows}
\usepackage{svg}
\pgfplotsset{compat=1.14}
\usetikzlibrary{calc}
\usetikzlibrary{patterns}
\usepackage[most]{tcolorbox}

\title{
On the Emotion Understanding of Synthesized Speech
}

\author{Yuan Ge$^{1}$\thanks{Equal contribution.}, Haishu Zhao$^{1*}$, Aokai Hao$^1$, Junxiang Zhang$^1$, Bei Li$^2$, Xiaoqian Liu$^1$, \\
\textbf{Chenglong Wang$^1$, Jianjin Wang$^1$, Bingsen Zhou$^1$, Bingyu Liu$^1$,} \\
\textbf{Jingbo Zhu$^{1,3}$, Zhengtao Yu$^4$, Tong Xiao$^{1,3}$}\thanks{Corresponding author.} \\
$^1$ Northeastern University, China 
$^2$ Meituan 
$^3$ NiuTrans Research \\ 
$^4$ Kunming University of Science and Technology
}

\begin{document}
\maketitle

\begin{abstract}
Emotion is a core paralinguistic feature in voice interaction. It is widely believed that emotion understanding models learn fundamental representations that transfer to synthesized speech, making emotion understanding results a plausible reward or evaluation metric for assessing emotional expressiveness in speech synthesis. In this work, we critically examine this assumption by systematically evaluating Speech Emotion Recognition (SER) on synthesized speech across datasets, discriminative and generative SER models, and diverse synthesis models.
We find that current SER models can not generalize to synthesized speech, largely because speech token prediction during synthesis induces a representation mismatch between synthesized and human speech. Moreover, generative Speech Language Models (SLMs) tend to infer emotion from textual semantics while ignoring paralinguistic cues. Overall, our findings suggest that existing SER models often exploit non-robust shortcuts rather than capturing fundamental features, and paralinguistic understanding in SLMs remains challenging.\footnote{\url{https://github.com/965002973/Synthesis_SER}}
\end{abstract}

\section{Introduction}

Speech understanding enables machines to extract meaning and intent from spoken signals, supporting robust interaction and reliable decision-making in real-world settings \citep{serdyuk2018towards, haghani2018audio, lugosch2019speech, borsos2023audiolm, wang2025speech}. 
Speech understanding is typically assumed to operate on naturally produced human speech, and the target information extends beyond semantic content to paralinguistic features and speaker identity. 
For example, the same expression `Really?' can convey distinct intents when spoken with surprise versus doubt emotions.

Speech understanding is increasingly challenged by the growing prevalence of synthesized speech \citep{xie2025towards, cui2025recent, hurst2024gpt}. As people can convey the same content via either recorded or synthesized audio, a natural question arises: Does understanding \textit{synthesized speech} differ from understanding human speech?

\definecolor{ugreen}{RGB}{098,190,166}
\definecolor{uyellow}{RGB}{253,186,107}
\definecolor{ured}{RGB}{235,096,070}
\definecolor{upurple}{RGB}{175,135,220} 
\definecolor{ublue}{RGB}{076,135,220}   
\definecolor{ugray}{RGB}{150,156,165}   
\definecolor{ubrown}{RGB}{176,125,086}  

\pgfmathsetlengthmacro{\BarW}{7pt}
\pgfmathsetlengthmacro{\BarStep}{9pt} 

\pgfmathsetlengthmacro{\ShiftHuman}{-3.5*\BarStep}
\pgfmathsetlengthmacro{\ShiftSynth}{-2.5*\BarStep}
\pgfmathsetlengthmacro{\ShiftCosy}{-1.5*\BarStep}
\pgfmathsetlengthmacro{\ShiftIndex}{-0.5*\BarStep}
\pgfmathsetlengthmacro{\ShiftKimi}{ 0.5*\BarStep}
\pgfmathsetlengthmacro{\ShiftGLM}{ 1.5*\BarStep}
\pgfmathsetlengthmacro{\ShiftTTS}{ 2.5*\BarStep}
\pgfmathsetlengthmacro{\ShiftAudio}{3.5*\BarStep}

\tikzset{
  solidbar/.style={},
  hatchbar/.style={
    postaction={draw=none, pattern=north east lines}
  }
}

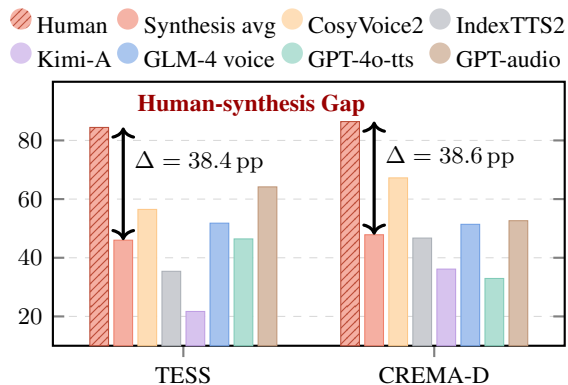
\begin{figure}[t]
    \centering
    \begin{tikzpicture}
    
        \begin{axis}[
            ybar,
            bar width=\BarW,
            width=1.08\linewidth,
            height=0.66\linewidth,
            ymin=10, ymax=100,
            ytick={20, 40, 60, 80},
            symbolic x coords={TESS,CREMA-D},
            xtick=data,
            enlarge x limits=0.52,
            axis line style={thick},
            tick style={thick},
            label style={font=\small},
            tick label style={font=\small},
            ymajorgrids,
            grid style={dashed, gray!30},
            xtick pos=left,
            legend style={
                font=\small,
                draw=none,
                fill=none,
                text opacity=1,
                rounded corners,
                at={(-0.1,1.3)},
                anchor=north west,
                legend cell align=left
            },
            legend image code/.code={
                \path[#1, draw=none, fill opacity=0.95] (0pt,-1pt) rectangle (8pt,7pt);
            },
            legend columns=4,
            clip=false
        ]
        
        \addplot+[
          solidbar,
          hatchbar,
          draw=ured!80!black,
          fill=ured!50,
          pattern color=ured!80!black,
          bar shift=\ShiftHuman
        ] coordinates {(TESS,84.43) (CREMA-D,86.43)};
        \addplot+[draw=ured!80, fill=ured!50, bar shift=\ShiftSynth] coordinates
            {(TESS,45.99) (CREMA-D,47.84)};
        \addplot+[draw=uyellow!80, fill=uyellow!50, bar shift=\ShiftCosy] coordinates
            {(TESS,56.48) (CREMA-D,67.23)};
        \addplot+[draw=ugray!80, fill=ugray!50, bar shift=\ShiftIndex] coordinates
            {(TESS,35.38) (CREMA-D,46.71)};
        \addplot+[draw=upurple!80, fill=upurple!50, bar shift=\ShiftKimi] coordinates
            {(TESS,21.69) (CREMA-D,36.13)};
        \addplot+[draw=ublue!80, fill=ublue!50, bar shift=\ShiftGLM] coordinates
            {(TESS,51.79) (CREMA-D,51.40)};
        \addplot+[draw=ugreen!80, fill=ugreen!50, bar shift=\ShiftTTS] coordinates
            {(TESS,46.42) (CREMA-D,32.94)};
        \addplot+[draw=ubrown!80, fill=ubrown!50, bar shift=\ShiftAudio] coordinates
            {(TESS,64.16) (CREMA-D,52.63)};
        
        \legend{Human, Synthesis avg, CosyVoice2, IndexTTS2, Kimi-A, GLM-4 voice, GPT-4o-tts, GPT-audio}
        
        \node[font=\small, text=red!60!black]
          at ([xshift=\ShiftSynth, xshift=48pt]axis cs:TESS,92)
          {\textbf{Human-synthesis Gap}};
        \draw[<->, very thick]
            ([xshift=\ShiftSynth]axis cs:TESS,45.99) --
            ([xshift=\ShiftSynth]axis cs:TESS,84.43)
            node[midway, yshift=8pt, right, font=\small] {$\Delta=38.4\,\mathrm{pp}$};
        
        \draw[<->, very thick]
            ([xshift=\ShiftSynth]axis cs:CREMA-D,47.84) --
            ([xshift=\ShiftSynth]axis cs:CREMA-D,86.43)
            node[midway, yshift=8pt, right, font=\small] {$\Delta=38.6\,\mathrm{pp}$};
        
        \end{axis}
    \end{tikzpicture}
    
    \caption {Speech emotion recognition accuracy on TESS and CREMA-D dataset. Synthetic speech results represent the agreement with human, as audio samples with indistinct emotional expressions were manually excluded. SER results utilizing Emotion2vec highlight a clear gap between human and synthesized speech.}
    \label{fig:intro}

\end{figure}

We investigate this shift through Speech Emotion Recognition (SER), a key component of voice-centric interaction. Prior work reports strong emotion recognition accuracy on human speech using both discriminative SER models and generative Speech Large Language Models (SLMs) \citep{koh2021comparison, sadok2023vector, ma2024emotion2vec, wu2025stepaudio2technicalreport}, yet generating speech that reliably expresses a target emotion remains challenging for text-to-speech (TTS) systems and speech-to-speech (S2S) LLMs \citep{du2024cosyvoice, chen2025f5, zeng2024glm, ding2025kimi}. 
Recent methods therefore evaluate or optimize synthesis using SER-based signals \citep{yang2025emovoice, an2024funaudiollm, wang2025spark, ji-etal-2025-controlspeech, chen2025emova, vevo, yang2025rlaif}. However, they assume that SER models learn emotion representations that transfer to synthesized speech, which has not been rigorously validated. 
This motivates our central research question: \textit{Can current models reliably understand emotion in synthesized speech?} 

To answer it, we systematically evaluate emotion understanding on human and synthesized speech across datasets, SER models (discriminative and generative), and synthesis paradigms (TTS and S2S LLMs). 
Our results provide four key findings:

\begin{itemize}
    \item Discriminative SER models generalize poorly under synthesized domain shift, and our findings highlight a clear gap between human and synthesized speech as shown in Fig. \ref{fig:intro}. 
    
    \item The dominant source of synthesized domain shift is the speech token prediction process, whereas the subsequent stages, 
    flow matching and vocoder, contribute substantially less.
    
    \item Supervised fine-tuning reduces the synthetic-domain gap, but neither standard fine-tuning nor domain-adversarial fine-tuning fundamentally resolves generalization, suggesting that SER models may rely on non-robust shortcuts.

    \item Speech LLMs tend to infer emotion primarily from textual semantic cues while ignoring paralinguistic signals. Prompt engineering does not change this behavior, indicating a persistent \textit{text dominance} \citep{wu2025language}.
\end{itemize}



\section{Related work}
\paragraph{Speech Emotion Recognition Model} The key problem to understand speech emotions is to learn the speech representation of the waveform \citep{baevski2020wav2vec, hsu2021hubert, chen2022wavlm, baevski2022data2vec} and finetune the models in downstream classification tasks \citep{koh2021comparison, sadok2023vector, ma2024emotion2vec}. We utilized Emotion2vec \citep{ma2024emotion2vec}, the most powerful and widely used open source SER model training on human speech, to evaluate the recognition performance on synthesized speech.
Furthermore, recent research emphasized the importance of evaluating emotional expressions in conversational contexts \citep{ge2025sagelm, zhao2026stylebench}.

\paragraph{Text-to-Speech Models} Emotional expression is critical in text-to-speech synthesized \citep{li2023styletts, du2024cosyvoice, yang2025emovoice, zhou2025indextts2}. 
However, we observe that the speech synthesized by most models fails to convey distinct emotions perceivable by human listeners.
We utilized CosyVoice 2 and IndexTTS 2, the open-source models with the most pronounced emotional expression, to generate TTS synthesized speech.

\paragraph{Speech-to-Speech LLMs} Emotion matters more in voice interaction between S2S LLMs and human \citep{hurst2024gpt, chu2024qwen2, zeng2024glm, wu2025stepaudio2technicalreport, xu2025qwen2, ding2025kimi}. We utilized Kimi-Audio and GLM4-Voice, the open-source S2S LLMs with the most pronounced emotional expression, to generate S2S synthesized speech.

\section{Preliminaries}

To validate whether current SER models reliably understand emotion in synthesized speech, we investigate their performance on three types of speech: real human speech, speech synthesized by TTS, and speech generated by S2S LLMs.

\begin{figure*}[t]
  \includegraphics[width=0.24\linewidth]{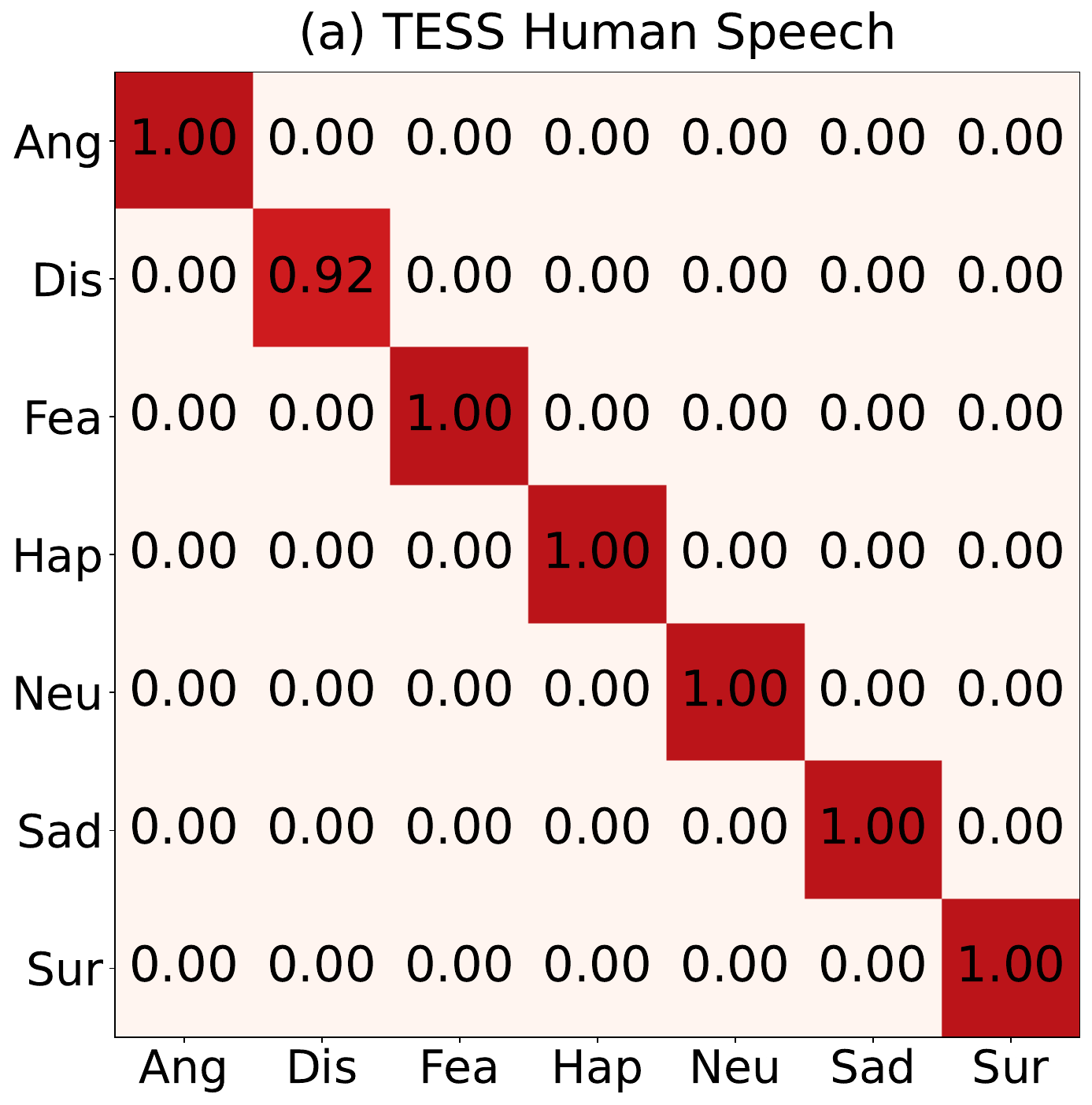} \hfill
  \includegraphics[width=0.24\linewidth]
  {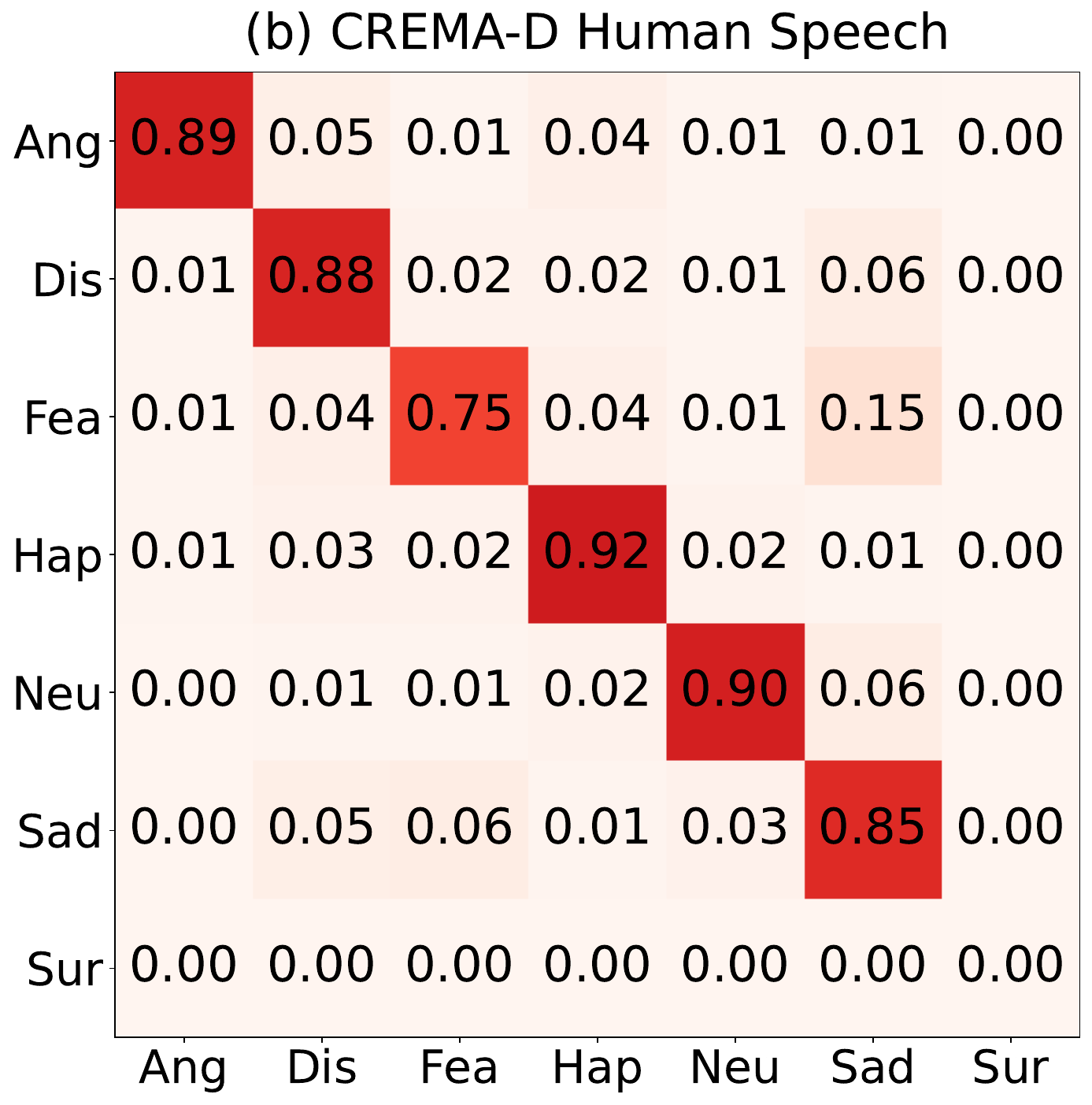} \hfill
  \includegraphics[width=0.24\linewidth]{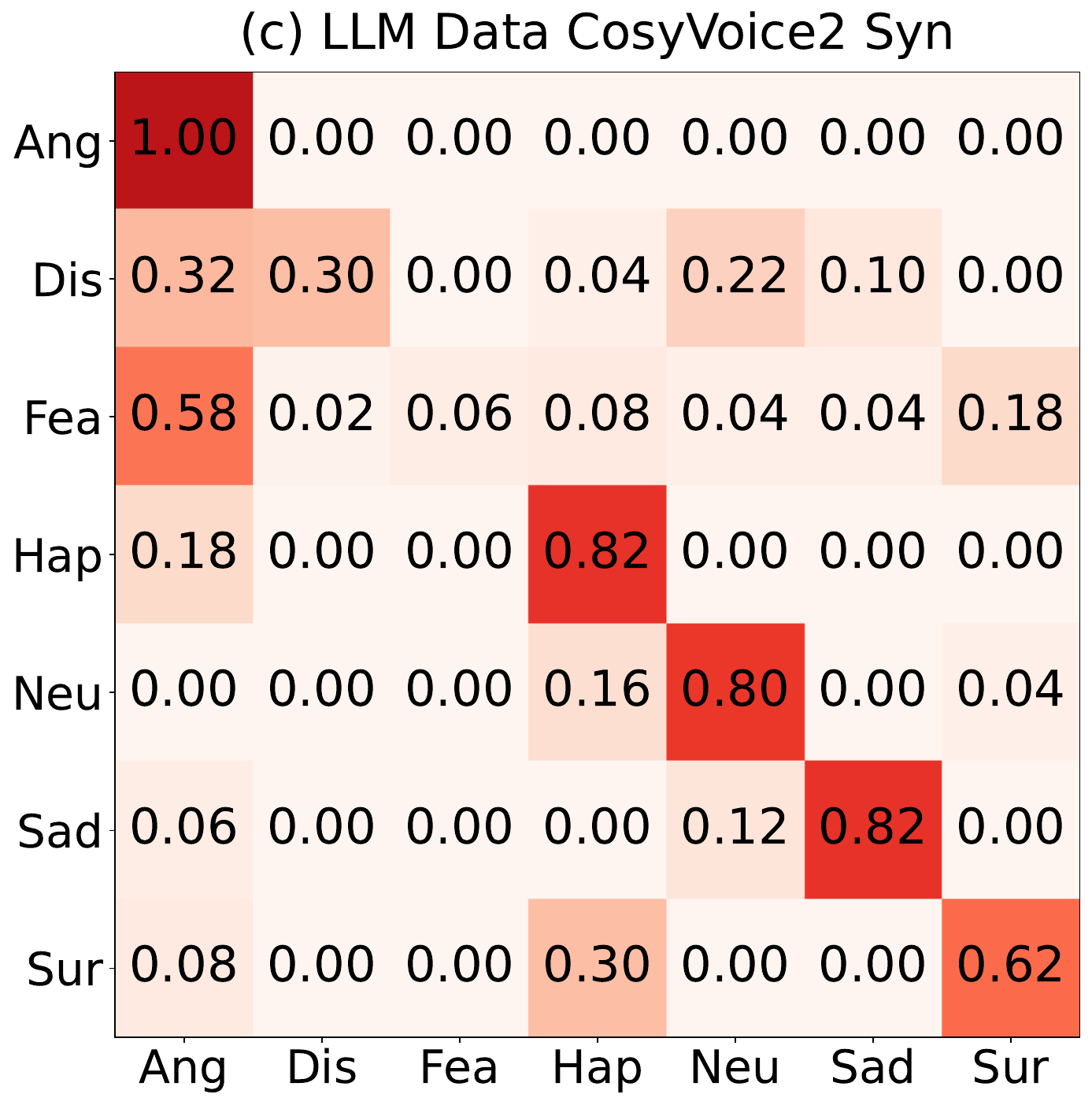} \hfill
  \includegraphics[width=0.24\linewidth]{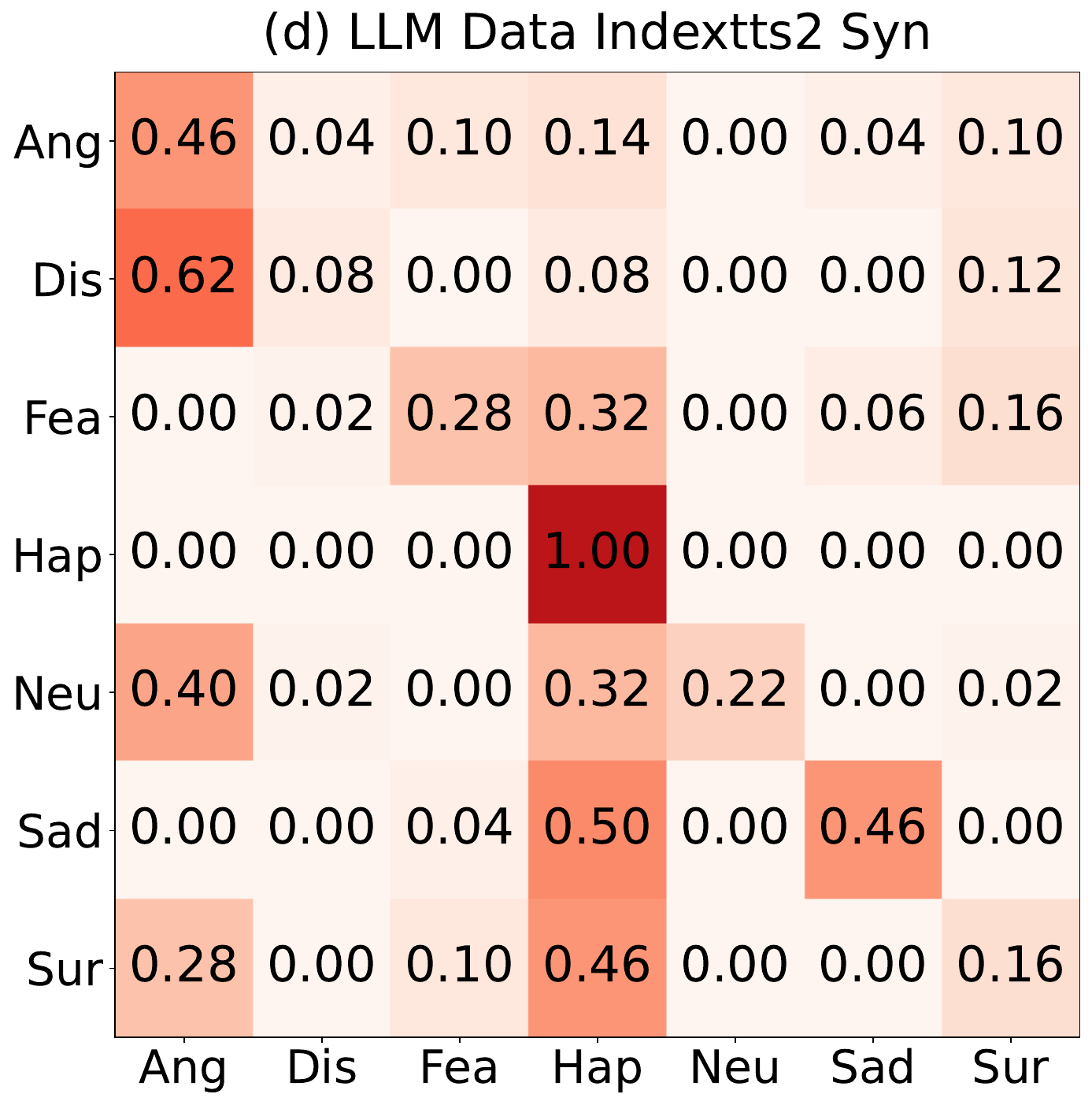} 
  
  \vspace{1ex} 
  \includegraphics[width=0.24\linewidth]{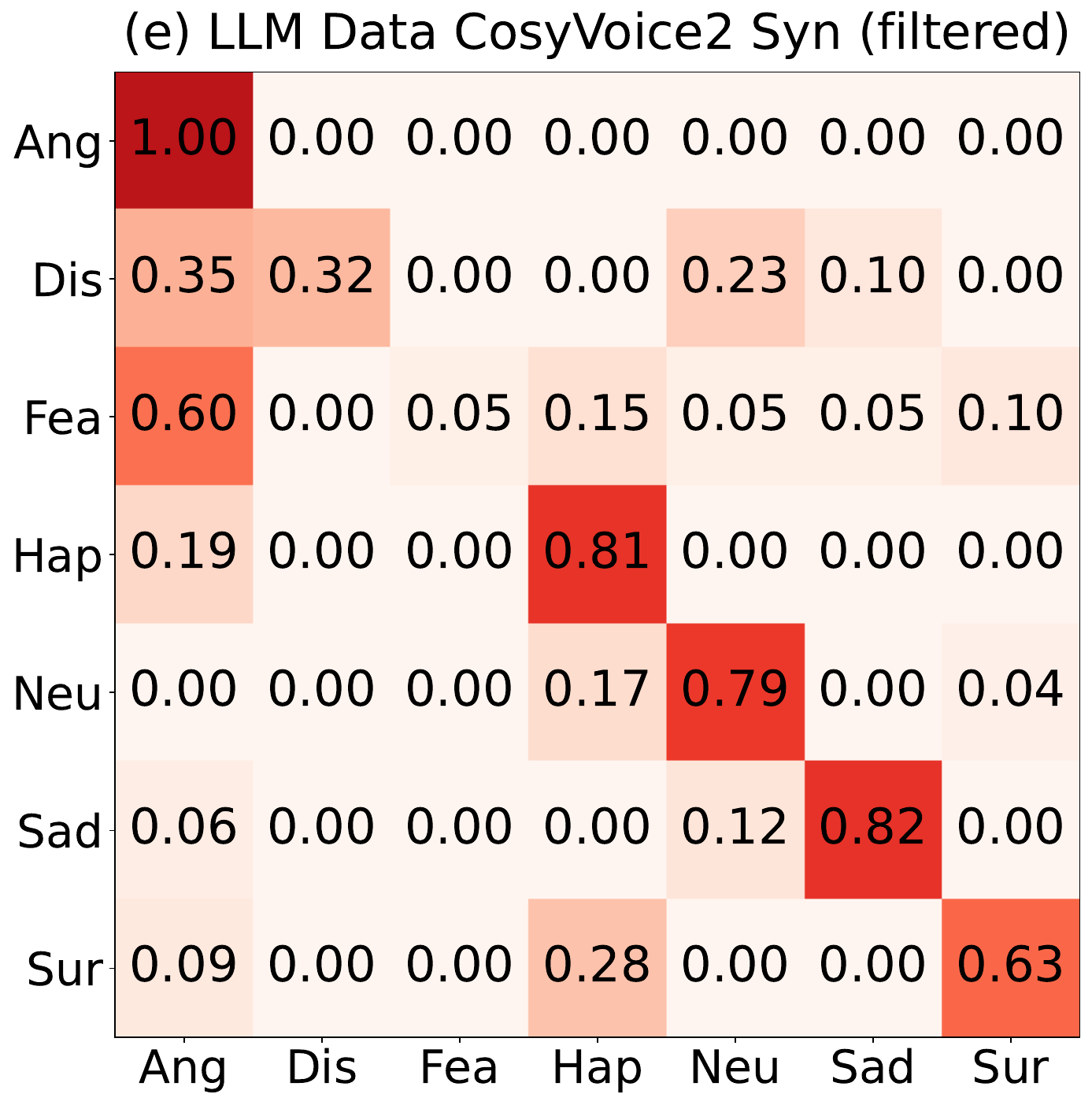} \hfill
  \includegraphics[width=0.24\linewidth]{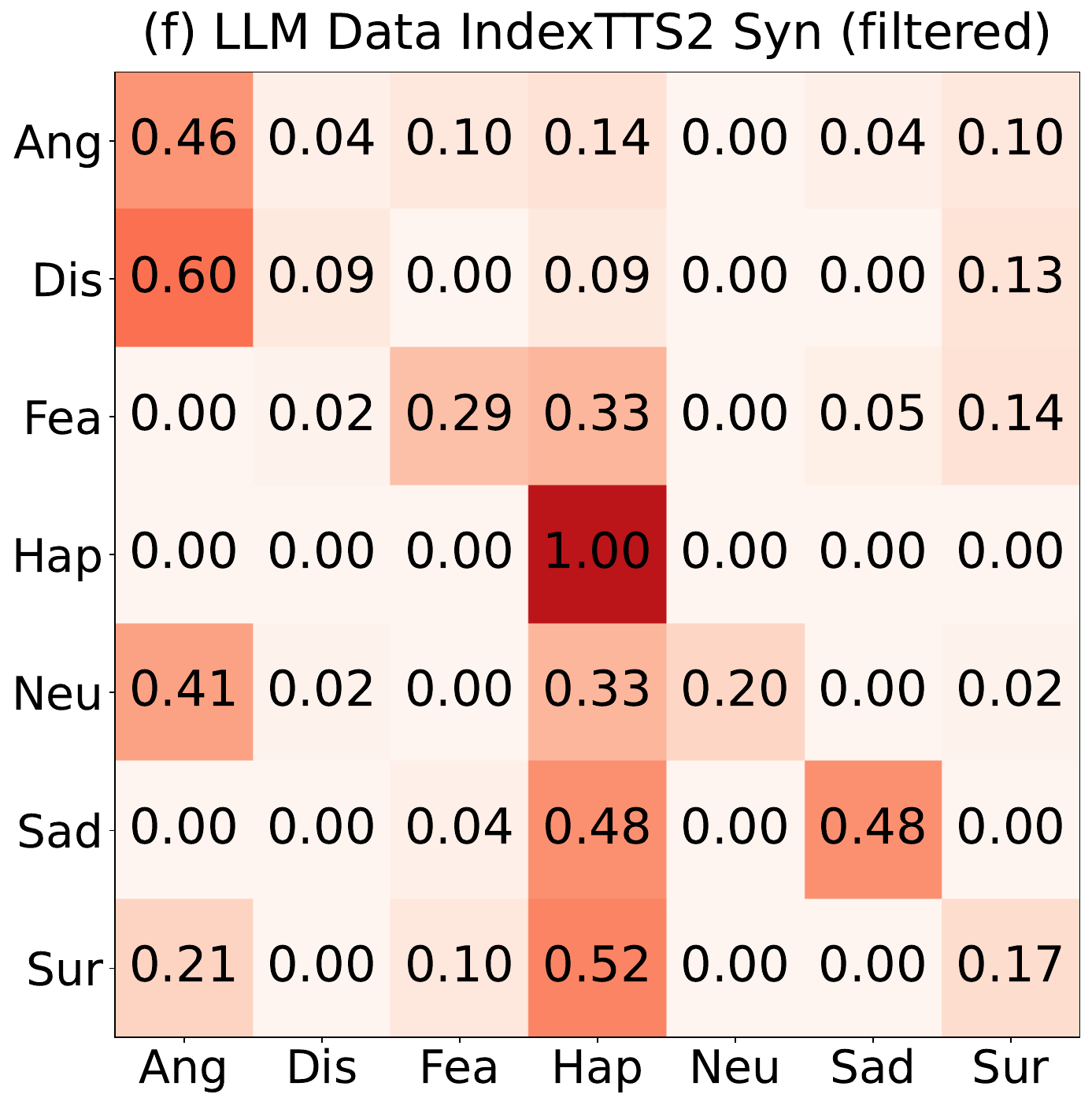} \hfill       
  \includegraphics[width=0.24\linewidth]{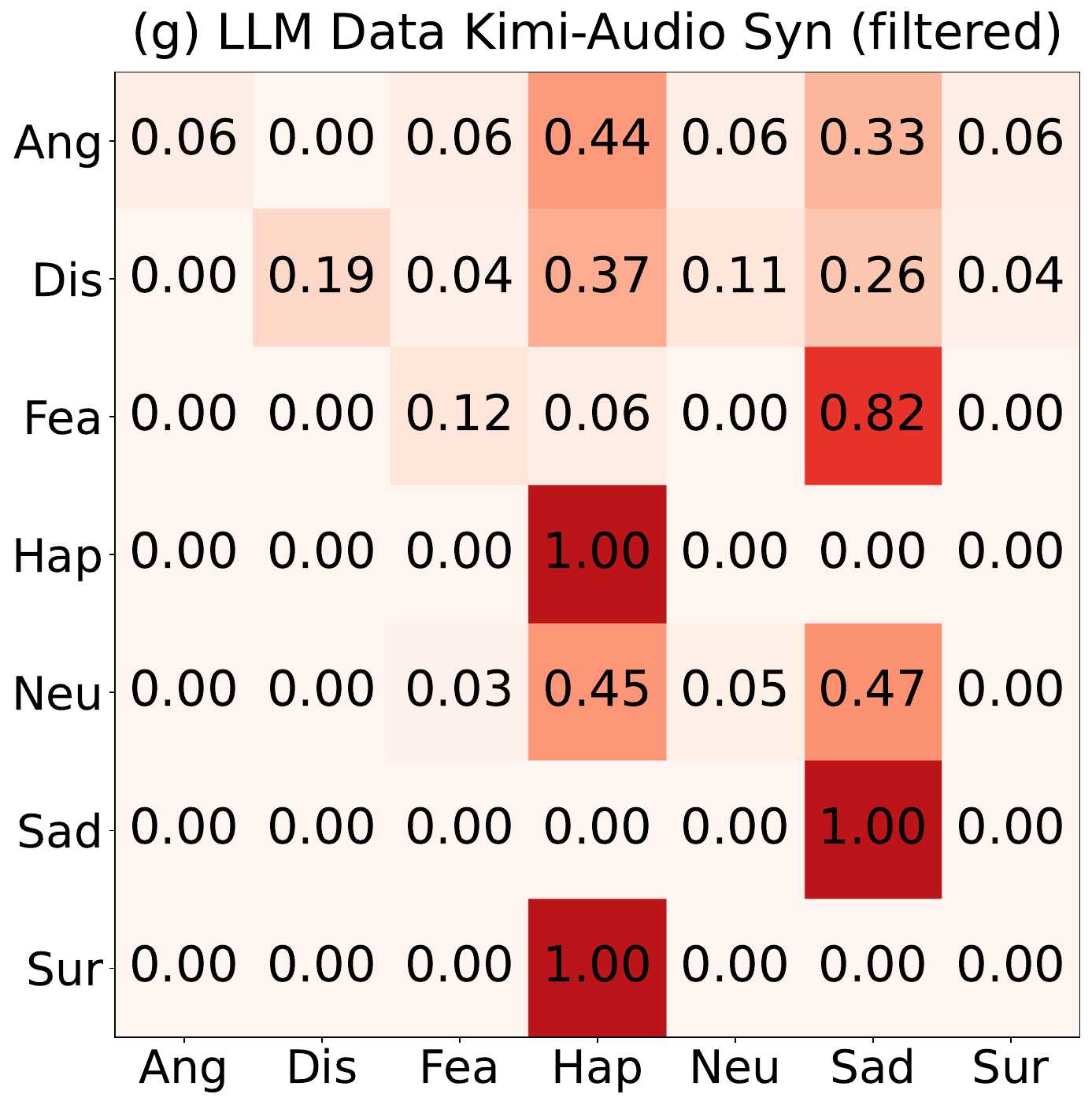} \hfill   
  \includegraphics[width=0.24\linewidth]{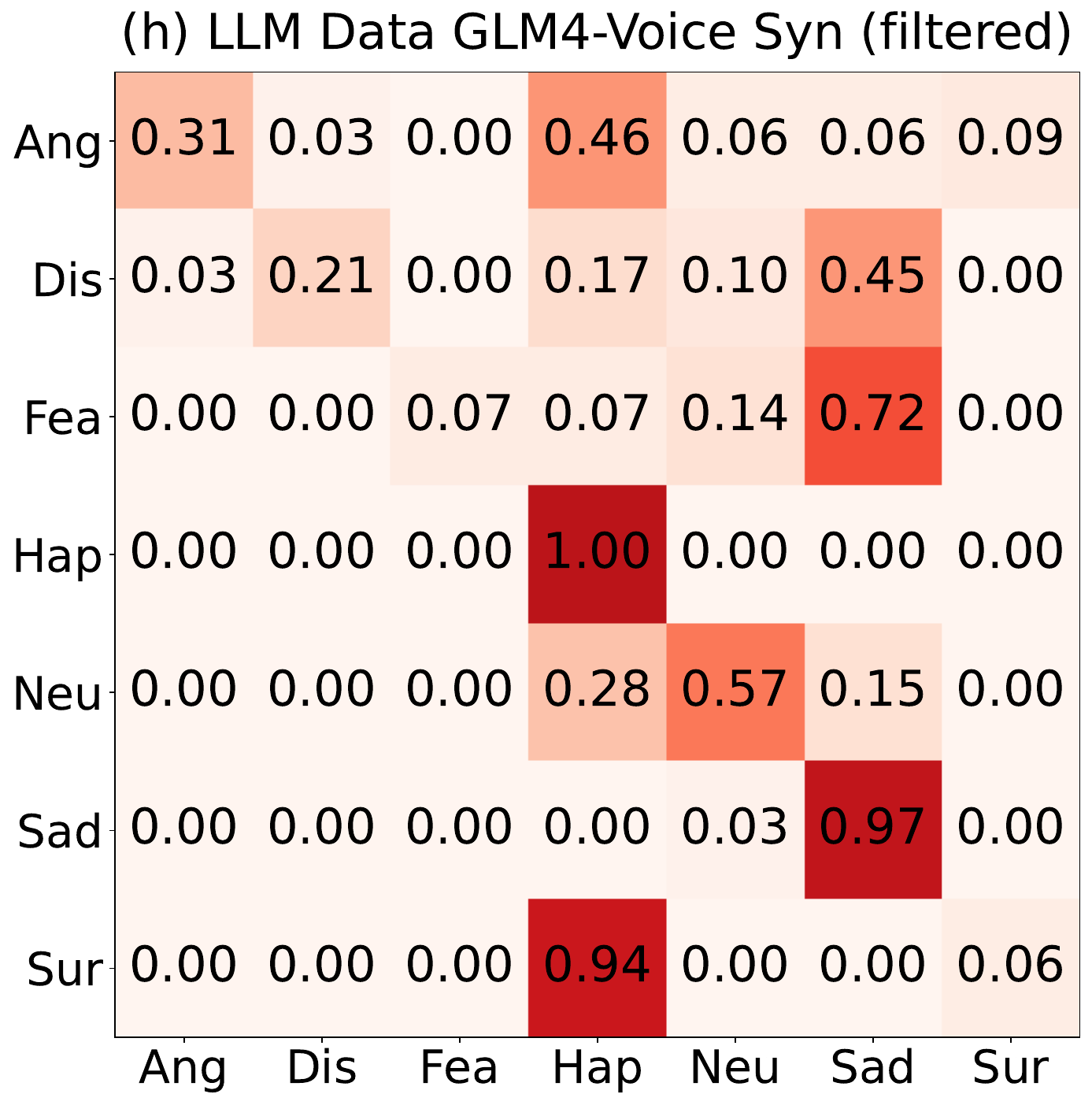}          

  \caption {The confusion matrix for speech emotion recognition is shown. The vertical axis represents the ground truth, and the horizontal axis represents the model's predictions. Sub-figures (a) and (b) show SER results on human speech, TESS and CREMA-D. Sub-figures (c) and (d) show SER results for speech synthesized by two TTS models from LLM-generated text. Sub-figures (e) and (f) represent the same results as (c) and (d) after filtering out weak emotional expression. Sub-figures (g) and (h) show the same as (e) and (f), but with synthesis by S2S LLMs.}
  \label{fig:preliminary}
\end{figure*}

\subsection{Task Formulation}

Formally, let $\mathcal{M}$ denote a SER model. The emotion recognition process can be formulated as:
\begin{equation}
E = \mathcal{M}(\mathcal{D}_S) 
\label{eq:SER}
\end{equation}
where $E$ represents the array of output emotion labels drawn from the discrete set \{angry, disgusted, fearful, happy, neutral, sad, surprised, other\}.
The input $\mathcal{D}_S$ denotes the speech dataset to be categorized, which may originate from natural human recordings, TTS models, or S2S LLMs. 
We formulate the speech synthesis process for TTS or S2S as:
\begin{equation}
\mathcal{D}_S = \mathcal{S}(\mathcal{D}_T, C) 
\label{eq:TTS}
\end{equation}
where $\mathcal{D}_T$ is the text dataset and $C$ is the emotion control signal. 
Despite the unified notation $C$, the realization of emotional control varies: TTS models typically utilize prompt speech cloning, natural instructions, or emotion embeddings to define emotions, whereas S2S LLMs require multi-turn interactions to achieve significant emotional output.
More details are available at Appendix \ref{appendix:s2s}.
Then the reverse process of synthesis is formulated as:
\begin{equation}
\mathcal{D}_T = \mathcal{A}(\mathcal{D}_S) 
\label{eq:ASR}
\end{equation}
where $\mathcal{A}$ represents Auto Speech Recognition (ASR), transcribing from speech distribution to text distribution using Whisper \citep{radford2023robust}.

\subsection{SER Results on Human Speech}
We investigate the performance of Emotion2vec in two test datasets: TESS \citep{TESS2020} and CREMA-D \citep{cao2014crema}. In the confusion matrix shown in Fig.~\ref{fig:preliminary} (a) and (b), the vertical axis represents the ground truth, while the horizontal axis represents the model's recognition results. Therefore, the sum of the values in each row equals 1.0. (However, note that because the `other' and `unknown' categories were excluded for conciseness, the actual sums may fall short of 1.0 in some instances.) Experimental results demonstrate that Emotion2vec achieves robust performance on human speech, yielding a confusion matrix with a pronounced diagonal structure.

\section{Domain Shift to Synthetic Speech}
In this section, we apply the discriminative SER model and generative SLMs to synthetic audio, examining the emotion understanding performance and analyzing the potential underlying causes.

\subsection{SER Results on Synthesized Speech}
Here, we investigate whether the SER model trained on human speech generalizes to synthesized speech. Specifically, we utilized gpt-4o to generate 3,028 text sentences and synthesize into speech using two TTS emotion control methods: prompt speech and natural instruction.

As shown in Fig. \ref{fig:preliminary} (c) and (d), the experimental results demonstrate that Emotion2vec achieves poor performance on TTS synthesized speech, producing a confusion matrix significantly without diagonal structure. Experimental results demonstrate that the SER model trained on human data performs well on human speech, while performs poorly on synthesized speech. So we explore the potential reasons for this performance gap.

\subsubsection{Hypothesis 1: TTS Emotion Expression}
As described in related work, we observed that the speech synthesized by most models fails to convey distinct emotions perceivable by human listeners. Despite employing TTS and S2S models with superior emotional expressiveness, certain synthesized utterances still fail to express the target emotion.

Consequently, we recruited four annotators to manually filter out synthesized utterances that failed to convey the target emotion. However, as illustrated in Fig. \ref{fig:preliminary} (e) and (f), the confusion matrix still lacks a distinct diagonal structure. Experimental results demonstrate that the SER model trained on human recorded data still generalizes poorly to the synthesized speech, even when the target emotions are clearly expressed.

\subsubsection{Hypothesis 2: Lack of Training Data}
Secondly, we examine whether the SER model exhibits performance degradation in specific emotion categories. Previous work by \citeauthor{yang2025emovoice} reported limited accuracy in Emotion2vec for disgusted, fearful, and surprised categories due to insufficient training data. Consistent with this, Fig. \ref{fig:preliminary} (e) confirm that the model struggles primarily with these three emotions other than neutral.
However, as shown in Fig. \ref{fig:preliminary} (f) (g) and (h), further experiments indicate a significant drop in performance across almost all emotion categories for synthesized speech generated by IndexTTS and S2S LLMs, including categories with sufficient training data, such as happy, angry, and sad. To ensure reliability, all synthesized audio employed in this section and hereafter was checked by human annotators to verify the distinctness of the emotional expression.

\begin{figure}[t]
  \centering
  \includegraphics[width=\columnwidth]{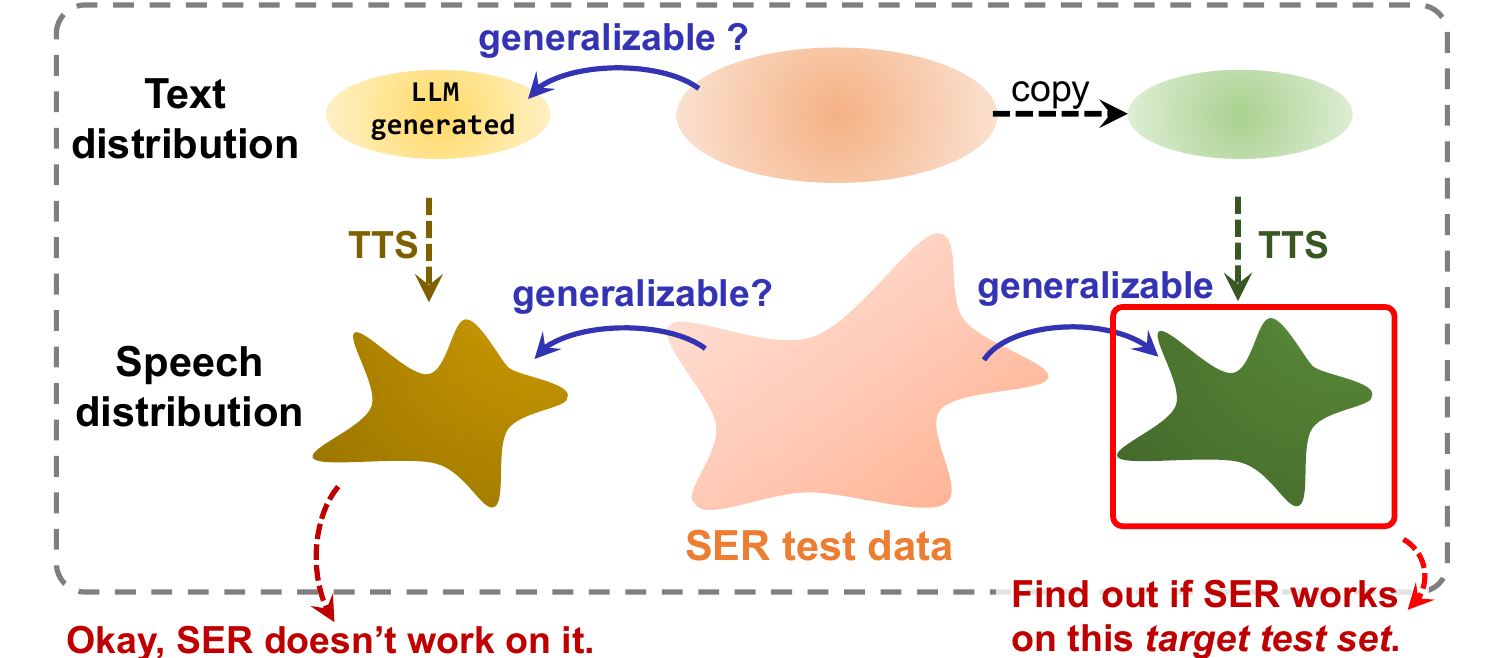}
  \caption{To mitigate the impact of text distribution generated by LLMs, we investigated speech emotion recognition performance on identical text datasets.}
  \label{fig:hyp3}
\end{figure}

\subsubsection{Hypothesis 3: Text Distribution Gap}
Moreover, we investigate to eliminate the concern that Emotion2vec lacks robustness to shifts in textual distribution. 
It is reasonable to hypothesize that the performance of the SER model is related to the underlying text distribution, where \textit{`text distribution'} is defined as the underlying semantic representation or ground truth ASR text sentences, identical with Eq. \ref{eq:ASR}.
As shown in Fig. \ref{fig:hyp3}, our initial recognition experiments on LLM-generated text (leftmost portion) yielded poor results. This raises a critical question: is the failure caused by an unfamiliar text distribution, or by the domain gap in the speech synthesis process?

To disentangle these factors, we conducted the ablation experiment depicted on the right side of Fig. \ref{fig:hyp3}. We first obtained ASR text transcripts from a test dataset, then synthesized text to generate a new target test set. The process to generate the target test dataset can be formulated as:
\begin{equation}
\mathcal{D}_S' = \mathcal{S}(\mathcal{A}(\mathcal{D}_S), C) 
\label{eq:target}
\end{equation}
where $\mathcal{S}(\cdot)$ represents speech synthesis and $\mathcal{A}(\cdot)$ denotes ASR.
This design ensures that any remaining performance drop can be attributed solely to the synthesis artifacts rather than the semantic content.

\begin{figure}[t]
  \includegraphics[width=0.48\linewidth]{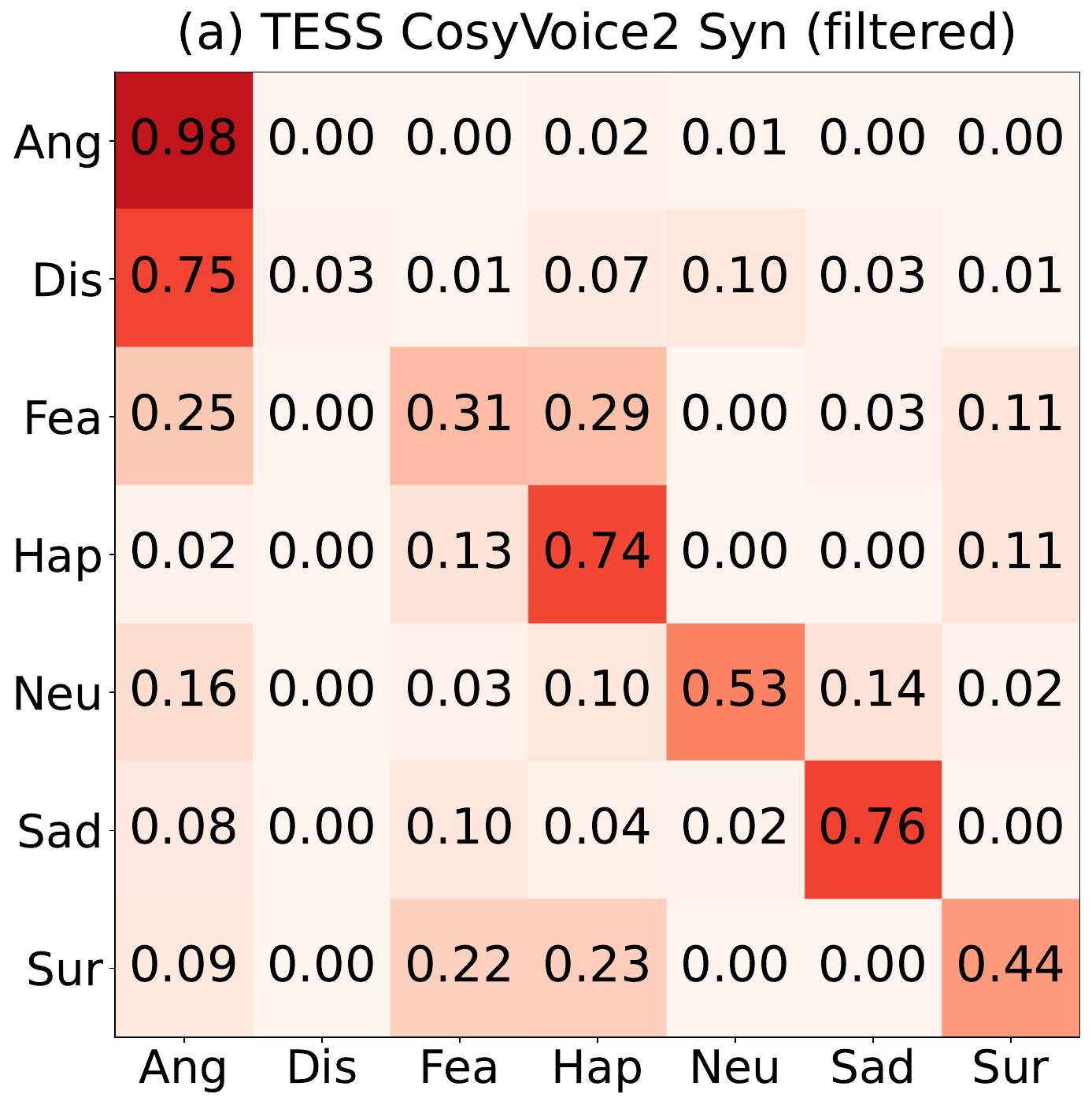} \hfill
  \includegraphics[width=0.48\linewidth]{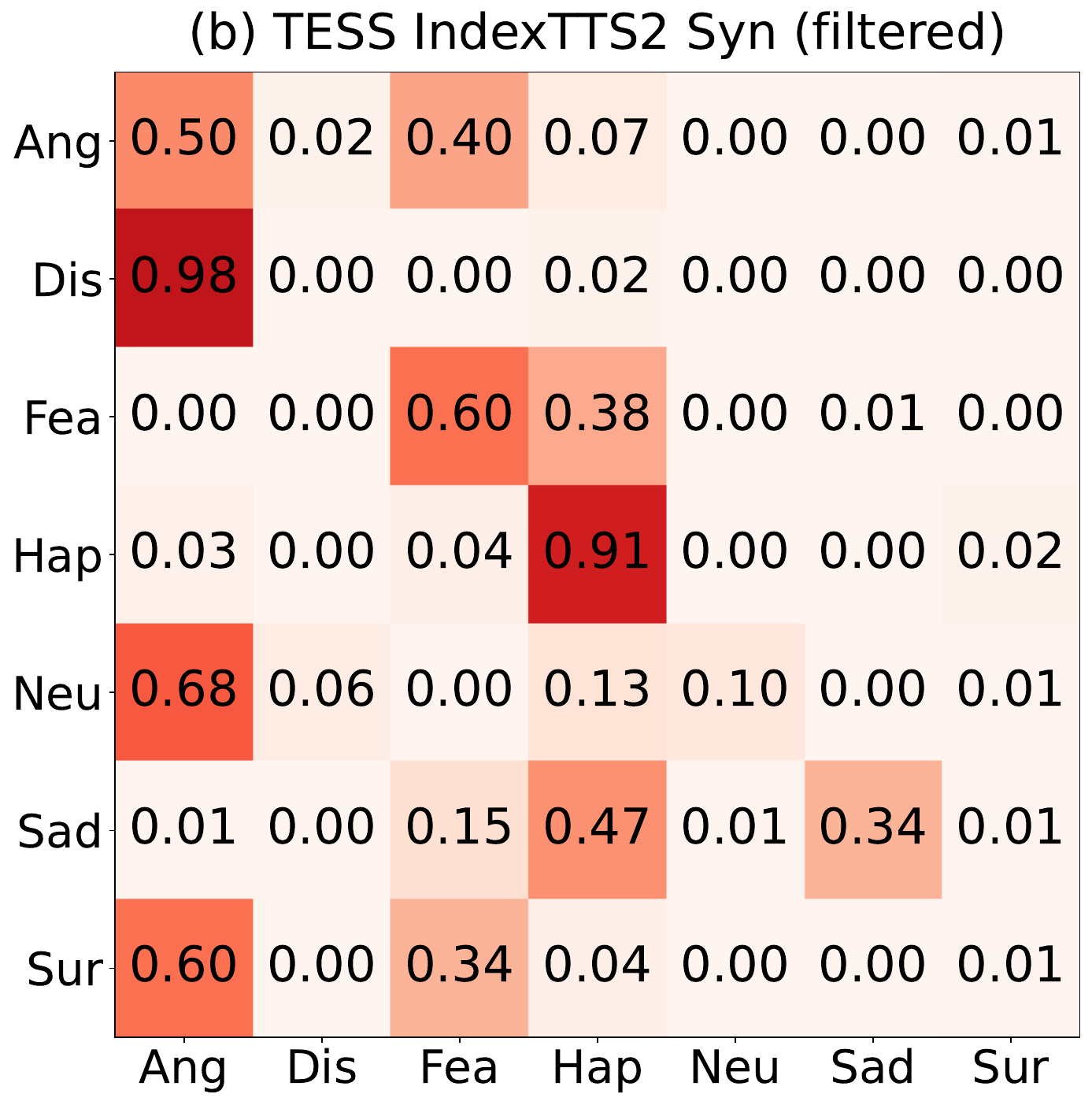}   
  \vspace{1ex}
  \includegraphics[width=0.48\linewidth]{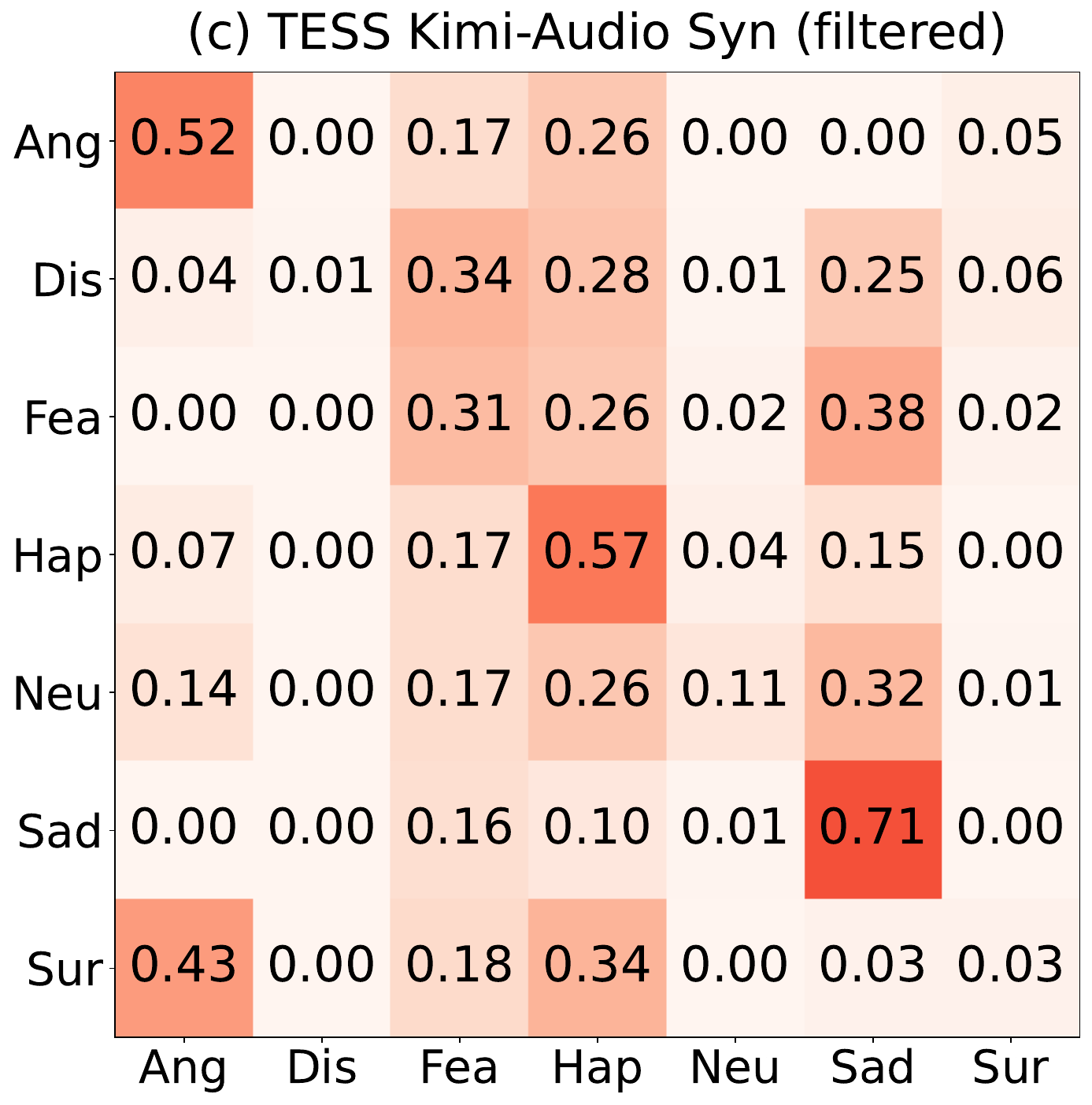} \hfill
  \includegraphics[width=0.48\linewidth]{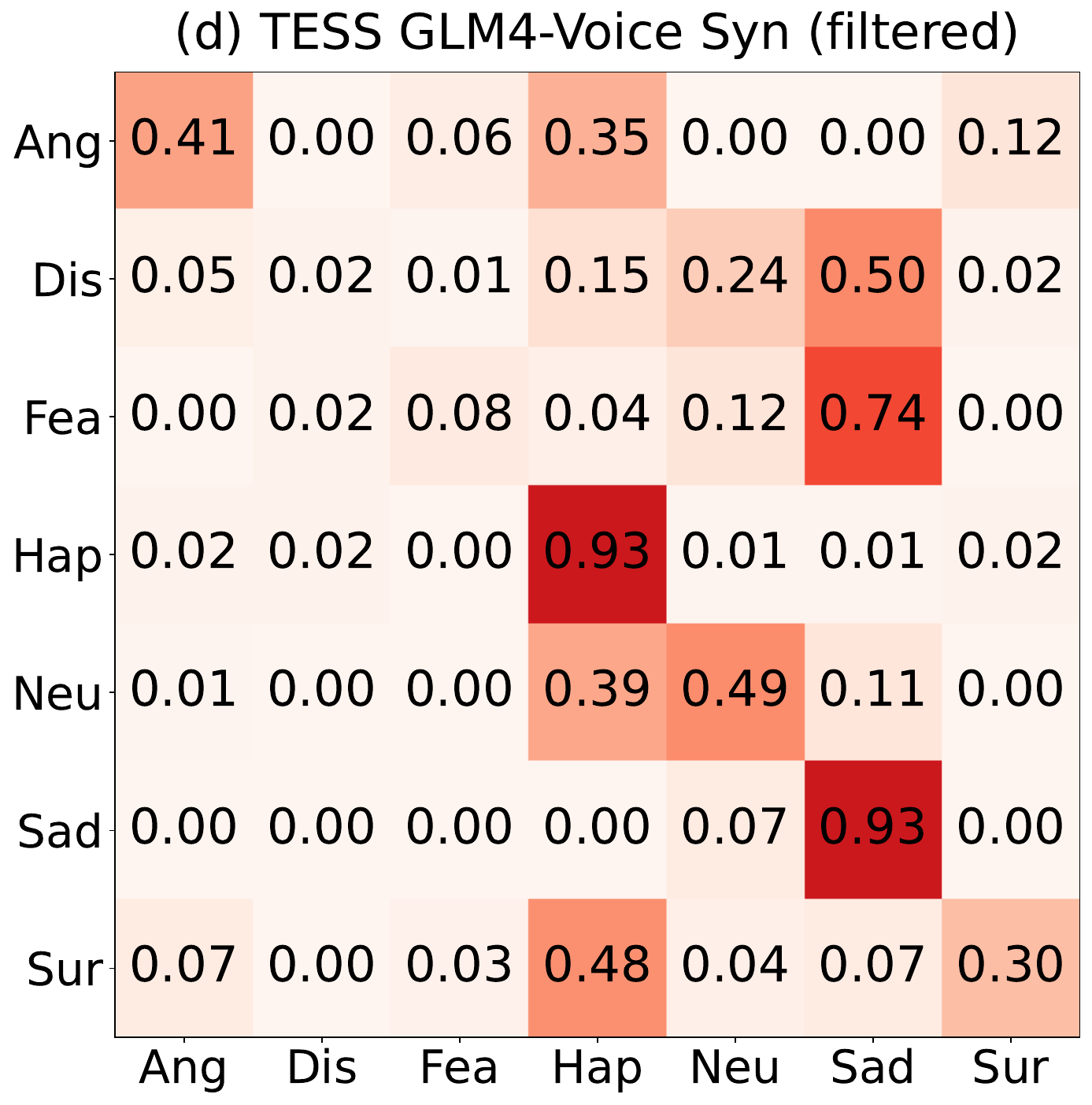} 
  

  \caption {Confusion matrix of speech emotion recognition results for synthetic speech based on TESS text. The vertical axis represents the ground truth, while the horizontal axis represents the model’s predictions. Sub-figs (a) and (b) show SER results for two TTS models, while (c) and (d) show SER results for two S2S models.}
  \label{fig:asr-tts}
\end{figure}

As shown in Fig. \ref{fig:asr-tts}, the confusion matrix of two TTS models and two S2S LLMs on synthesized TESS dataset continues to exhibit a lack of diagonal structure. This suggests that performance degradation of the SER model on synthesized speech does not stem from a text distribution gap. Instead, it likely arises from synthesis artifacts introduced by the speech generation process. 
All synthesized audio employed in this section and below was synthesized from a real human speech transcript.

\begin{table}[htbp]
    \centering
    \resizebox{\linewidth}{!}{
    \begin{tabular}{lccccc}
        \toprule
         & \multicolumn{3}{c}{\textbf{Human}} & \multicolumn{2}{c}{\textbf{Synthesized}} \\
        \cmidrule(lr){2-4} \cmidrule(lr){5-6}
         & IEM. & RAV. & TESS & IEM. & TESS \\
        \midrule
        \multicolumn{6}{l}{\cellcolor{black!10} \textbf{\textit{I. Emotion2vec Train on Human Speech}}} \\
        Ang & 77.23 & 81.25 & 97.50 & 63.27 & 50.00  \\
        Hap & 64.14 & 56.25 & 100.00 & 80.30 & 0.00 \\
        Neu & 71.98 & 100.00 & 100.00 & 43.24 & 0.00 \\
        Sad & 75.68 & 50.00 & 100.00 & 24.59 & 0.00 \\
        Fea & - & 93.75 & 100.00 & - & 55.00 \\
        Dis & - & 81.25 & 100.00 & - & 0.00 \\
        Sur & - & 93.75 & 100.00 & - & 3.57 \\
        \midrule
        WA & 71.28 & 77.88 & 99.64 & 49.58 & 15.31 \\
        \midrule
        \multicolumn{6}{l}{\cellcolor{black!10} \textbf{\textit{II. Emotion2vec Train on Synthesized Speech}}} \\
        Ang & 48.51 & 25.00 & 0.00 & 85.71 & 97.22 \\
        Hap & 16.67 & 0.00 & 0.00 & 74.24 & 70.59 \\
        Neu & 71.98 & 37.50 & 100.00 & 96.22 & 94.59 \\
        Sad & 87.16 & 100.00 & 2.50 & 86.89 & 92.59 \\
        Fea & - & 0.00 & 0.00 & - & 85.00 \\
        Dis & - & 0.00 & 2.50 & - & 96.77 \\
        Sur & - & 0.00 & 0.00 & - & 92.86 \\
        \midrule
        WA & 55.67 & 22.12 & 15.00 & 89.20 & 91.84 \\
        \bottomrule
    \end{tabular}
    }
    \caption{SER accuracy on real or synthesized speech while trained with human or synthesized speech data. The IEMOCAP dataset contains only 4 emotion labels.}
    \label{tab:gap}
\end{table}

\subsubsection{Hypothesis 4: Distribution Gap between Synthesized and Human Speech}
Prior experiments suggest that a SER model trained on human speech does not generalize to understanding tasks on synthesized audio. Even when ASR transcripts are identical, conveying the same semantics and content, there remains a mismatch between the audio distributions produced by natural speech production and by the speech synthesis process.

To further validate the distribution gap between synthesized and human speech, we fine-tune a pretrained audio representation model, Emotion2vec-base \citep{ma2024emotion2vec}, using either human or synthesized data under controlled settings. Specifically, for real speech we use three datasets: IEMOCAP, RAVDESS, and TESS. We then generate a synthesized counterpart by first transcribing the real utterances with ASR, performing emotion-conditioned speech synthesis from the transcripts, and filtering out samples with insufficiently perceivable emotional expression. Because RAVDESS contains substantial transcript duplication, the resulting synthesized set is relatively small; therefore, all synthesized RAVDESS samples are used for training, without holding out a test split.

As shown in Table \ref{tab:gap}, models trained on real speech perform well on real audio but degrade sharply on synthesized audio. Conversely, models trained on synthesized audio classify synthesized utterances more accurately yet generalize poorly to real speech. These results confirm a pronounced distribution gap between synthesized and human speech, and highlight the limitation of directly reusing SER models trained on real speech for synthesized audio understanding.
\
\subsection{Commercial Synthesis Models}

SER models trained on human audio exhibit significant performance degradation on synthetic audio, even the emotional expression of the synthesized audio is manually verified. 
Given that open-source models frequently utilize inaccurate labels by Emotion2vec as RL training rewards or evaluation metrics, we question whether this phenomenon persists in leading commercial models and whether this discrepancy contributes to the performance gap between open-source and commercial models. 

To this end, we employ the TTS model \texttt{GPT-4o mini TTS} and the S2S model \texttt{GPT-4o Audio} to synthesize natural, realistic, and emotionally expressive speech, subsequently evaluating the SER models performance. The results show that the gap persists despite stronger affective expressiveness: on CREMA-D, Emotion2vec achieves 32.94\% accuracy on gpt-4o-tts outputs and 52.63\% on gpt-4o-audio outputs. On the more lexically homogeneous TESS dataset, accuracy improves but remains low at 46.42\% and 64.16\%, respectively. More detailed results are provided in Appendix \ref{appendix:commercal}. 
These findings indicate that even for leading commercial synthesizers, current human-trained SER models do not reliably recognize emotion in synthesized speech.

\subsection{SER Results with SLMs}

\definecolor{uyellow}{RGB}{253,186,107}
\definecolor{ured}{RGB}{235,096,070}
\definecolor{upurple}{RGB}{175,135,220} 
\definecolor{ublue}{RGB}{076,135,220}  

\pgfmathsetlengthmacro{\BarW}{8pt}
\pgfmathsetlengthmacro{\BarStep}{10pt} 

\pgfmathsetlengthmacro{\ShiftQwenA}{-1.5*\BarStep}
\pgfmathsetlengthmacro{\ShiftGptA}{  -0.5*\BarStep}
\pgfmathsetlengthmacro{\ShiftQwenB}{  0.5*\BarStep}
\pgfmathsetlengthmacro{\ShiftGptB}{   1.5*\BarStep}

\tikzset{
  solidbar/.style={},
  hatchbar/.style={
    postaction={draw=none, pattern=north east lines}
  }
}

\begin{figure}[t]
    \centering
    \begin{tikzpicture}
    \tikzstyle{textonly} = [font=\scriptsize,align=left]
    \tikzstyle{sublayers} = [rectangle,draw,minimum width=0.4cm,rounded corners=0pt,align=center,inner sep=0pt,minimum height=0.2cm,font=\scriptsize]
    \begin{axis}[
        ybar,
        bar width=\BarW,
        width=1.05\linewidth,
        height=0.62\linewidth,
        ymin=0, ymax=90,
        ytick={0,20,40,60,80},
        symbolic x coords={SLM,real,syn},
        xtick={SLM,real,syn},
        xticklabels={SLM Datasets, Human, Synthesis},
        enlarge x limits=0.32,
        axis line style={thick},
        tick style={thick},
        label style={font=\normalsize},
        tick label style={font=\normalsize},
        ymajorgrids,
        grid style={dashed, gray!30},
        xtick pos=left,
        clip=false
    ]

    \addplot+[draw=ured!80, fill=ured!50, bar shift=\ShiftQwenA] coordinates
        {(SLM,69.23)};

    \addplot+[draw=uyellow!80, fill=uyellow!50, bar shift=\ShiftQwenB] coordinates
        {(SLM,76.58)};

    \addplot+[draw=ublue!80, fill=ublue!50, bar shift=\ShiftQwenA] coordinates
        {(real,18.87) (syn,18.87)};

    \addplot+[draw=upurple!80, fill=upurple!50, bar shift=\ShiftQwenB] coordinates
        {(real,27.1) (syn,27.68)};   
    
    \addplot+[solidbar, hatchbar, draw=ured!80, fill=ured!50, pattern color=ured!80!black, bar shift=\ShiftGptA] coordinates
        {(SLM,66.67)};
    
    \addplot+[solidbar, hatchbar, draw=uyellow!80, fill=uyellow!50, pattern color=uyellow!80!black, bar shift=\ShiftGptB] coordinates
        {(SLM,80.67)};

    \addplot+[solidbar, hatchbar, draw=ublue!80, fill=ublue!50, pattern color=ublue!80!black, bar shift=\ShiftGptA] coordinates
        {(real,11.56) (syn,11.56)};
    
    \addplot+[solidbar, hatchbar, draw=upurple!80, fill=upurple!50, pattern color=upurple!80!black, bar shift=\ShiftGptB] coordinates
        {(real,20.86) (syn,26.59)};

        
    \end{axis}
    \node[sublayers, color=black, fill=black!2, fill opacity=1](tip) at (2.5,2.6) {};
    \node[textonly] at ([xshift=0.9cm]tip) {Qwen3-Omni};
    \node[sublayers, color=black, fill=black!2, pattern=north east lines, pattern color=black] at ([xshift=1.8cm]tip) {};
    \node[textonly] at ([xshift=2.9cm]tip) {GPT-4o Audio};

    \node[sublayers, color=ured, fill=ured!50, fill opacity=1] at ([yshift=-0.3cm]tip) {};
    \node[textonly] at ([xshift=0.9cm, yshift=-0.3cm]tip) {StepEval};
    \node[sublayers, color=uyellow, fill=uyellow!50, fill opacity=1] at ([xshift=1.8cm, yshift=-0.3cm]tip) {};
    \node[textonly] at ([xshift=2.9cm, yshift=-0.3cm]tip) {Multi-speaker ES};
    \node[sublayers, color=ublue, fill=ublue!50, fill opacity=1] at ([yshift=-0.6cm]tip) {};
    \node[textonly] at ([xshift=0.9cm, yshift=-0.6cm]tip) {TESS};
    \node[sublayers, color=upurple, fill=upurple!50, fill opacity=1] at ([xshift=1.8cm, yshift=-0.6cm]tip) {};
    \node[textonly] at ([xshift=2.9cm, yshift=-0.6cm]tip) {CREMA-D};
    
    \end{tikzpicture}
    
    \caption {Speech emotion recognition accuracy of SLMs. Solid colors and shaded areas represent the SER results of Qwen3-omni and GPT-4o Audio respectively.}
    \label{fig:SLM}

\end{figure}
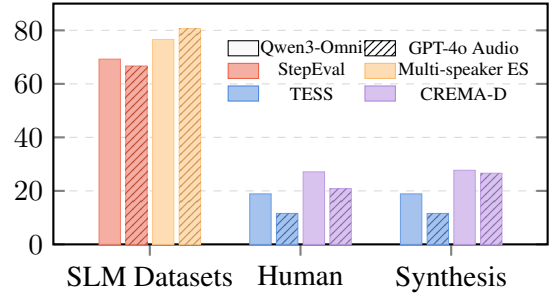

SLMs are central to voice-centric interaction and are believed to have strong speech and emotion understanding capabilities. This section explores the performance of advanced SLMs on both human and synthesized speech datasets. We selected two SLMs including the open-source Qwen3-Omni \citep{xu2025qwen3omnitechnicalreport} and the commercial GPT-4o Audio. In addition to TESS and CREMA-D, we also include two datasets used to assess the emotion understanding of SLMs: StepEval-Audio-Paralinguistic \citep{wu2025step} and the Multi-Speaker Emotional Speech Dataset\footnote{\url{https://magichub.com/datasets/multi-speaker-emotional-speech-dataset/}}.

As shown in Fig. \ref{fig:SLM}, both models achieve high SER accuracy on the two SLM datasets, with an average of 73.3\%. However, on TESS and CREMA-D, emotion recognition accuracy drops significantly, averaging only 19.6\%. The models also perform poorly on synthesized data.

Interestingly, this phenomenon suggests that emotion understanding of SLMs relies more on semantic content than on paralinguistic cues. In TESS and CREMA-D, the same text is read with different emotions, requiring emotion inference from audio's paralinguistic features. In contrast, audio in StepEval and Multi-speaker SE can typically be interpreted based on the transcribed text. Furthermore, prompt engineering aimed at directing SLMs to focus more on paralinguistic information does not have an effect (specific prompts are provided in the Appendix \ref{appendix:prompt}). This highlights the `text dominance' in SLMs \citep{wu2025language} and suggests the need for further enhancement of their ability to process paralinguistic features.

\subsection{Top-tier SER and S2S LLMs}

\begin{table}[t]
    \centering
    \resizebox{\linewidth}{!}{
    \begin{tabular}{lccccc}
        \toprule
         & \multicolumn{3}{c}{\textbf{Human}} & \multicolumn{2}{c}{\textbf{Synthesized}} \\
        \cmidrule(lr){2-4} \cmidrule(lr){5-6}
         & IEM. & RAV. & TESS & IEM. & TESS \\
        \midrule
        Emotion2vec & 71.28 & 77.88 & 99.64 & 49.58 & 15.31 \\
        C2SER & 78.64 & 53.85 & 47.50 & 74.24 & 54.59 \\
        Gemini 2.5 Pro & 65.10 & 30.77 & 22.14 & 72.02 & 26.53 \\
        \bottomrule
    \end{tabular}
    }
    \caption{Emotion recognition accuracy of top-tier open-source SER model and commercial S2S LLM.}
    \label{tab:top-tier}
\end{table}

\begin{figure}[t]
  \includegraphics[width=0.48\linewidth]{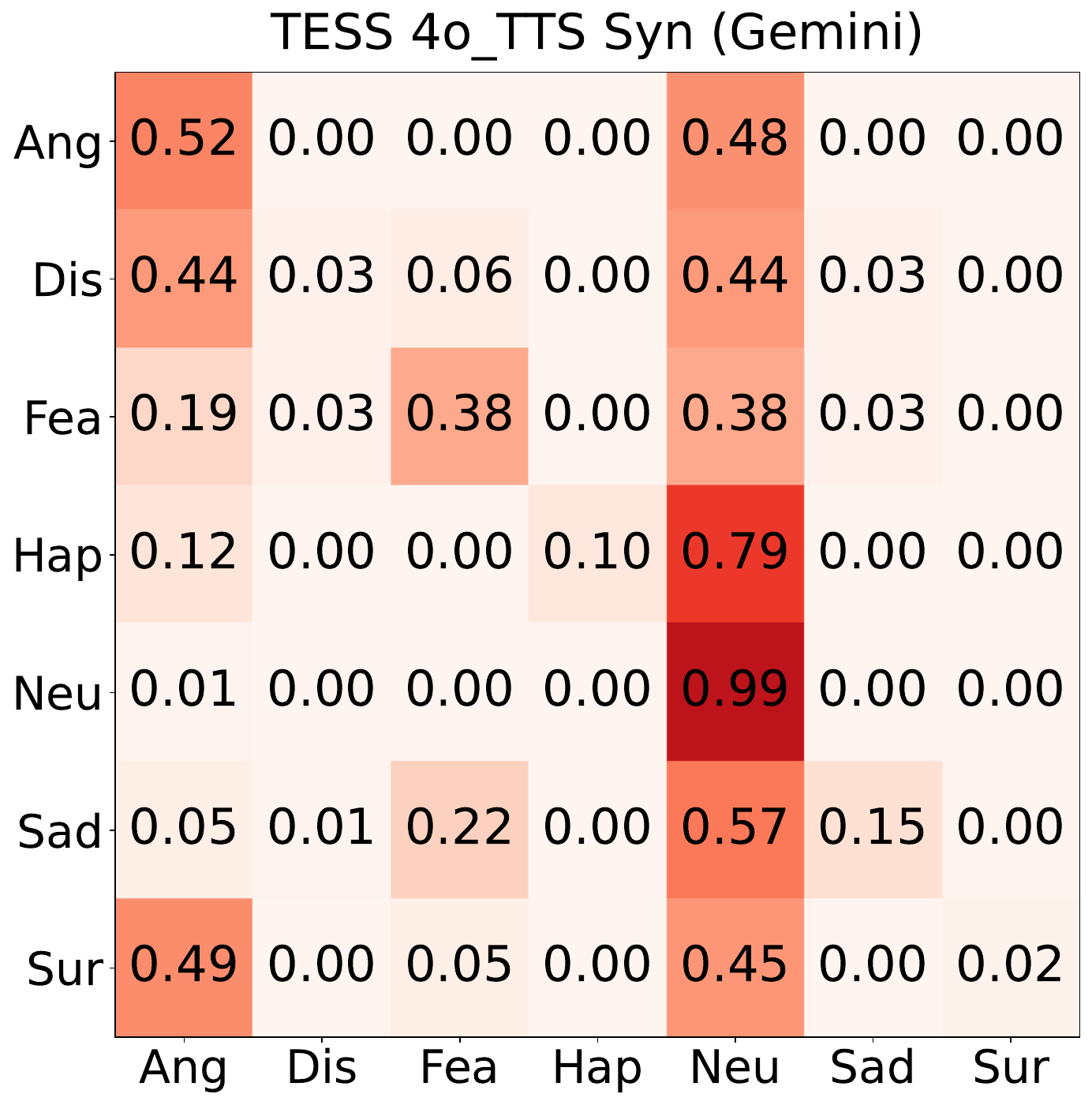} \hfill
  \includegraphics[width=0.48\linewidth]{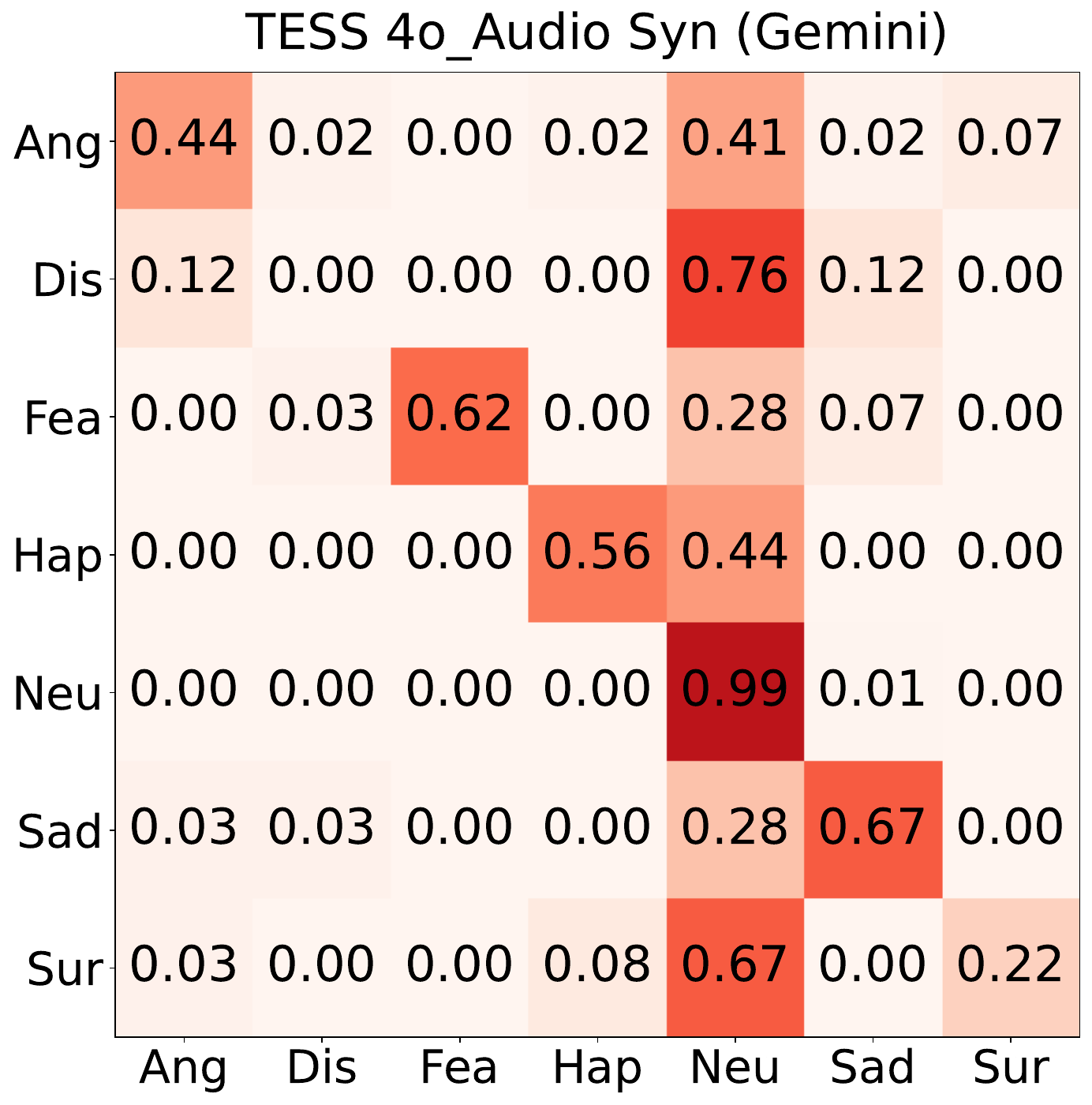} 

  \caption {Confusion matrix of the speech emotion recognition results of Google Gemini 2.5 pro. All samples are synthesized by GPT-4o TTS and GPT-4o Audio.}
  \label{fig:gemini-results}
\end{figure}

SER performance on synthesis speech could be influenced by (a) the capacity of the SER model and (b) the capacity of the synthesis models to generate intended emotional expression. We investigate the performance of top-tier SER models utilizing open-source C2SER \cite{zhao2025steering} and commercial Gemini 2.5 Pro.
As shown in Table \ref{tab:top-tier}, even these advanced models continue to exhibit a significant human-synthesis gap, demonstrating that this remains a robust challenge across the current landscape of SER and S2S models. 
Additionally, to understand whether top-tier SER models can accurately recognize emotions from top-tier TTS models, we evaluated Gemini on speech synthesized by GPT-4o-mini-tts and GPT-4o-audio. As shown in Fig. \ref{fig:gemini-results}, our results indicate that even when both the TTS and SER are top-tier commercial models, the accuracy of emotion recognition remains suboptimal. This confirms that the gap is not merely a limitation of weaker models but a fundamental issue in synthesized emotional speech.

\begin{figure*}[t]
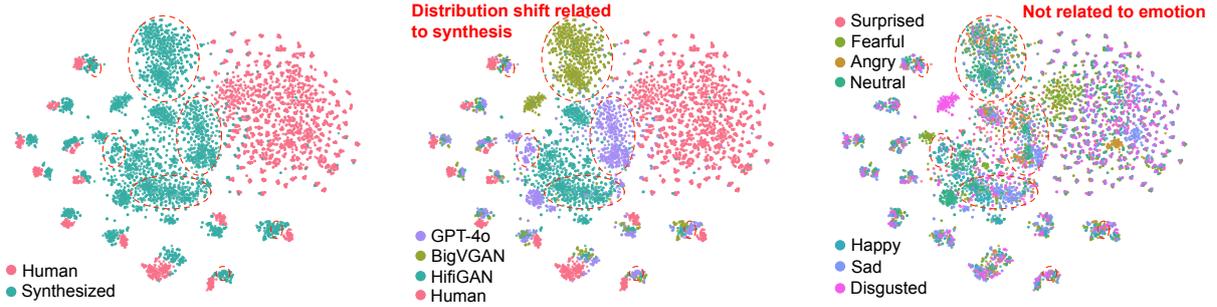

  \includegraphics[width=0.32\linewidth]{figs/test_synthesis.pdf} \hfill
  \includegraphics[width=0.32\linewidth]{figs/test_vocoder.pdf} \hfill
  \includegraphics[width=0.32\linewidth]{figs/test_emotion.pdf} 

  \caption {Visualization of the Emotion2vec embedding space, distinguished by synthesis type, vocoder, and emotion.}
  \label{fig:visualize}
\end{figure*}

\section{Deep Analysis}

\subsection{Representation Space Bias}

To investigate the significant performance drop of Emotion2vec on synthesized audio, we hypothesize that it is related to the model's representation space. The pretraining objective of Emotion2vec focuses on distilling speech representations from the teacher model, data2vec, via self-supervised learning. However, the pretraining data does not include synthesized audio, and data2vec \citep{baevski2023efficient} does not model synthetic data well. Consequently, a gap in the representation space may exist between synthesized and real speech, causing fine-tuned classification heads to be more influenced by distributional differences in the synthesized audio rather than emotional features.

\paragraph{Representation space shift of synthetic audio.} We begin by visualizing the representation space of Emotion2vec \textbf{utilize t-SNE} \citep{maaten2008visualizing}. Specifically, we encode real and synthesized data from the TESS and CREMA-D datasets, then reduce the representation to two dimensions using t-SNE. Three coloring schemes are used to distinguish samples: whether the data are synthesized, the audio's vocoder, and emotion.

Fig.~\ref{fig:visualize} supports our hypothesis. Subfigure 1 shows that the clusters of synthesized data are offset from those of real data, marked by red dashed circles. These offsets are attributed to specific synthesis models, such as CosyVoices 2 with the HiFiGAN vocoder at the bottom and IndexTTS 2 with the BigVGAN vocoder in other clusters. Subfigure 3 reveals that these outlier clusters encompass all emotions, indicating that the offsets are model-specific rather than emotion-related. Thus, classification models trained in this representation space are likely influenced by differences between synthesis models, hindering the extraction of emotion.

\begin{table}[t]
    \centering
    \small
    \begin{tabular}{lcccc}
    \toprule[0.7pt]
        \textbf{Metric(\%)} & \textbf{Syn.} & \textbf{Voc.}  & \textbf{Mod.} & \textbf{Emo.}\\ \hline
        \specialrule{0em}{1pt}{1pt}
        Balanced Acc & 99.81 & 96.25 & 75.01 & 59.86 \\ 
        Macro-F1 & 99.81 & 96.25 & 76.74 & 59.19 \\ 
    \toprule[0.7pt]
    \end{tabular}
    \caption{Linear probing results on Emotion2vec representation space. Target probing categories including synthesis / human (Syn.), vocoder name (Voc.), model name (Mol.), and emotion label (Emo.).}
    \label{tab:probe}
\end{table}

\paragraph{Embeddings capture more synthesis-specific patterns.} Probing experiments further corroborate this finding. Using a dataset balanced across emotion labels and synthesis models, we attached linear probes to frozen Emotion2vec embeddings to predict four targets: authenticity, vocoder, synthesis model, and emotion. As shown in Table \ref{tab:probe}, the probes easily discriminate between real and synthesized audio, vocoders, and even synthesis models, but perform poorly on emotion classification. These results motivate us to reshape the Emotion2vec representation space to disentangle synthesis-related artifacts from emotional features. 

\subsection{Failure of Generalization}

\begin{table*}[htbp]
    \centering
    \resizebox{\textwidth}{!}{
    \begin{tabular}{lcccccccccc}
        \toprule
        & \multicolumn{5}{c}{\textbf{Human Speech}} & \multicolumn{4}{c}{\textbf{Synthesized Speech}} \\
        \cmidrule(lr){2-6} \cmidrule(lr){7-10}
        \multicolumn{1}{c}{\textbf{Method}} & \multicolumn{5}{c}{\textbf{In Domain}} & \multicolumn{2}{c}{\textbf{In Domain}} & \multicolumn{2}{c}{\textbf{OOD}} \\
        \cmidrule(lr){2-6} \cmidrule(lr){7-8} \cmidrule(lr){9-10}
        & IEMOCAP & RAVDESS & TESS & MELD & CREMA-D & IEMOCAP & TESS & CREMA-D $\dagger$ & CREMA-D $\ddagger$ \\
        \midrule
        MLP & \textbf{71.13} & 66.35 & \textbf{99.64} & \textbf{50.15} & \textbf{78.43} & \textbf{89.20} & 88.27 & 32.77 & 49.38 \\
        DANN\_syn & 69.22 & \textbf{70.19} & 98.21 & 47.85 & 72.10 & 84.21 & 84.18 & 47.30 & \textbf{50.21} \\ 
        DANN\_vocoder & 69.07 & 69.23 & 97.14 & 47.24 & 71.70 & 84.49 & 85.20 & \textbf{51.35} & 46.88 \\ 
        DANN\_model & 69.37 & 68.27 & 98.21 & 47.97 & 71.96 & 81.99 & 83.16 & 41.22 & 43.87 \\ 
        DANN\_syn* & 67.60 & 69.23 & 98.93 & 47.85 & 73.16 & 86.43 & \textbf{91.33} & 42.91 & 45.01 \\ 
        DANN\_vocoder* & 70.54 & 65.38 & 97.86 & 47.12 & 72.10 & 84.21 & 88.27 & 44.93 & 46.78 \\ 
        DANN\_model* & 69.81 & 69.23 & 98.21 & 49.00 & 71.43 & 85.04 & 89.80 & 46.28 & 47.51 \\ 
        \bottomrule
    \end{tabular}
    }
    \caption{Under different fine-tuning strategies, the SER models fail to demonstrate generalization, both to cases where training data includes real CREMA-D and speech synthesis by CosyVoice and Kimi-audio but lacks CREMA-D synthesis with these models (OOD CREMA-D $\dagger$), and to unseen synthesis models (OOD CREMA-D $\ddagger$). The asterisk (*) indicates that positive and negative samples are balanced within each mini-batch.}
    \label{tab:generalize}
\end{table*}

\input{figs/method}

Table \ref{tab:gap} reveals a pronounced distribution gap between synthesized and human speech. Meanwhile, labeled synthetic speech is scarce: although fine-tuning uses 40k labeled human speech, only 4k synthetic speech remain after manual validation and filtering of weak emotional expression. A natural approach is to fine-tune on a mixture of both domains, since Table \ref{tab:gap} suggests that large-scale human speech pre-training combined with a small amount of labeled synthetic data can yield strong synthesis SER performance, as illustrated in Tabel \ref{tab:gap}.
Moreover, biases in the Emotion2vec representation space motivate disentangling synthesis-related artifacts from emotional cues. Inspired by domain adversarial neural networks \citep{ganin2016domain}, we learn latent representations that support the emotion classifier while confusing a domain classifier.

\paragraph{Domain adversarial fine-tuning.} As shown in Fig. \ref{fig:method}, 
the input text is first processed by a pretrained Emotion2vec, and we keep the pretrained parameters frozen to prevent catastrophic forgetting. The output embeddings are then fed into a trainable MLP Feature Extractor. This module projects the fixed embeddings into a task-specific latent representation, denoted as $h$, which serves as the shared feature space for subsequent tasks.
Then an emotion classifier is optimized to predict the correct emotion label $y$ by minimizing the cross-entropy loss $\mathcal{L}_{emo}$. To achieve disentanglement, we train an adversarial classifier so that the learned representation is invariant to the domain label $d$.
In practice, we implement this using a Gradient Reversal Layer (GRL,  \citealp{ganin2015unsupervised}), which negates the gradients flowing from the domain classifier to the MLP during backpropagation. 
The overall training objective is to learn a representation $h$ that is discriminative for emotion recognition but invariant to domain shifts.

In experimental settings, the human speech data comprise IEMOCAP, RAVDESS, TESS, MELD, and CREMA-D. For synthesized speech, we transcribe IEMOCAP and TESS into text and synthesize audio using CosyVoice2 and Kimi-Audio. All data are split into training and test sets. The test set additionally includes two out-of-domain (OOD) partitions, CREMA-D$\dagger$ and CREMA-D$\ddagger$, to assess generalization. Notably, training does not include synthesized CREMA-D, but it does include human CREMA-D and synthesized audio from other datasets produced by CosyVoice2 and Kimi-Audio; thus, CREMA-D$\dagger$ evaluates the \textit{compositional generalization}. CREMA-D$\ddagger$ evaluates generalization to unseen synthesis models, including IndexTTS2, GLM4-voice, gpt-4o-tts, and gpt-4o-audio. The training data for the following experimental results consist of both real and synthetic audio.

\paragraph{Generalization limits of fine-tuned SER.} As shown in Table \ref{tab:generalize}, simply training an MLP on mixed-domain data yields high in-domain accuracy, averaging 73.14\% on human speech and 88.73\% on synthesized speech, yet generalization remains poor. Performance drops sharply on OOD CREMA-D$\dagger$ and CREMA-D$\ddagger$.
Although we adopt domain-adversarial training with a GRL to prevent the latent representation $h$ from encoding synthesis attributes, and the rising domain loss $\mathcal{L}_{\text{domain}}$ suggests that this objective is being enforced. However, contrary to our expectation, this does not improve OOD generalization for SER.
Balancing positive and negative samples within each mini-batch also fails to help. These results suggest that SER models still exploit in-domain shortcuts rather than learning fundamental emotion features. 
Further experimental results regarding the fine-tuning of LALMs for SER are provided in Appendix \ref{appendix:lalms}.

\subsection{Token Prediction Dominates the Bias}

Furthermore, we investigate the contribution of each stage in the speech synthesis process to the distributional bias.
Recent speech synthesis \citep{xie2025towards, cui2025recent} typically consists of three stages: (1) an autoregressive (AR) language model generating discrete speech tokens; (2) a flow matching process synthesizing mel-spectrograms conditioned on the speech tokens and a reference audio; and (3) a vocoder converting mel-spectrograms into waveforms. To investigate the impact of each stage on SER accuracy, we conducted an ablation study using CosyVoice 2 on the TESS dataset.
We isolated the error contribution of each stage as follows:
\begin{itemize}
    \item Vocoder Error: We reconstructed audio from ground-truth mel-spectrograms. The performance drop compared to real speech represents the vocoder-induced error.
    
    \item Flow Matching Error: We synthesized mel-spectrograms using ground-truth speech tokens derived by the tokenizer as conditions. The performance gap compared to reconstruction from mel-spectrograms reflects the error introduced by flow matching process.
    
    \item Speech Tokens Generation Errors: We generated speech tokens from ASR-transcribed text, and synthesis speech with the following two stages. The discrepancy with the ground-truth token setup isolates the error introduced by the autoregressive language model.
\end{itemize}

\definecolor{acadBlue}{RGB}{110, 138, 175}
\definecolor{acadRed}{RGB}{168, 56, 54}
\definecolor{acadGreen}{RGB}{138, 176, 54}
\definecolor{acadBrown}{RGB}{196, 126, 68}
\definecolor{darkGrey}{RGB}{80, 80, 80}

\begin{figure}[t]
\centering
\begin{tikzpicture}
    \tikzset{
        errorbrace/.style={
            decorate,
            decoration={brace, amplitude=5pt, raise=2pt}, 
            thick, 
            black 
        }
    }

    \begin{axis}[
    width=1.12\linewidth, 
    height=6.5cm,
    grid=major,
    grid style={dashed, gray!30},
    xlabel style={font=\small, yshift=-3pt},
    ylabel style={font=\small},
    xtick={1,2,3},
    xticklabels={Text, Speech Token, Mel Spectrogram},
    xticklabel style={font=\footnotesize},
    tick label style={font=\footnotesize},
    ymin=25, ymax=90,
    ytick={25,40,60,80},
    axis lines=box,
    axis line style={line width=1pt},          
    every axis plot/.append style={             
        line width=1.5pt,
        mark size=2.5pt
    },
    legend style={
        at={(0.66,0.32)},
        anchor=north west,
        draw=none,
        fill=white,
        fill opacity=0.8,
        text opacity=1,
        font=\small,
        legend columns=1,
        column sep=2pt,
        nodes={scale=0.8, transform shape}
    },
    ]

    \addplot[color=darkGrey, mark=diamond*, mark options={fill=white}] coordinates {
        (1, 34.57) (2, 84.07) (3, 82.43)
    };
    \addlegendentry{Self}

    \addplot[color=acadBlue, mark=*, mark options={fill=white}] coordinates {
        (1, 32.79) (2, 83.93) (3, 82.43)
    };
    \addlegendentry{Truncation}

    \addplot[color=acadBrown, mark=triangle*, mark options={fill=white}] coordinates {
        (1, 30.21) (2, 81.43) (3, 82.43)
    };
    \addlegendentry{Speaker}

    \addplot[color=acadGreen, mark=square*, mark options={fill=white}] coordinates {
        (1, 29.14) (2, 76.29) (3, 82.43)
    };
    \addlegendentry{Dataset}

    \addplot[very thick, red, domain=1:3.3, samples=2] {83.93} 
        node[below, at start, align=left, xshift=18pt, font=\bfseries\small, red] {Human Speech\\Accuracy};

    \coordinate (D_Token) at (axis cs:2, 76.29);   
    \coordinate (B_Token) at (axis cs:1, 29.14); 
    \coordinate (M_Token) at (axis cs:2, 29.14); 
    
    \coordinate (D_Mel)   at (axis cs:3, 82.43);
    \coordinate (B_Mel)   at (axis cs:2, 76.29);
    \coordinate (M_Mel) at (axis cs:3, 76.29); 

    \coordinate (D_Human) at (axis cs:3.3, 83.93);
    \coordinate (B_Human) at (axis cs:3, 82.43);
    \coordinate (M_Human) at (axis cs:3.3, 82.43);

    \end{axis}

    \draw[thick, dashed, black!80] (B_Token) -- ++(2.5, 0);
    \draw[errorbrace] ($(D_Token)+(0,-0.1)$) -- (M_Token) 
        node[midway, right=6pt, align=left, font=\bfseries\small, color=acadRed] 
        {Speech Tokens \\ Generation Errors};

    \draw[thick, dashed, black!80] (B_Mel) -- ++(2.5, 0);
    \draw[errorbrace] ($(D_Mel)+(0,-0.1)$) -- (M_Mel) 
        node[midway, below=4pt, xshift=-10pt, align=left, font=\bfseries\small, color=acadRed] 
        {Flow Matching Errors};

    \draw[thick, dashed, black!80] (B_Human) -- ++(0.8, 0);
    \draw[errorbrace] (D_Human) -- (M_Human)
        node[midway, above=2pt, xshift=-16pt, align=left, font=\bfseries\small, color=acadRed] 
        {Vocoder Errors};
        
    \draw[->, >=stealth, ultra thick, acadRed] (D_Token) -- (M_Token);
    \draw[->, >=stealth, ultra thick, acadRed] (D_Mel) -- (M_Mel);
    \draw[->, >=stealth, ultra thick, acadRed] (D_Human) -- (M_Human);

\end{tikzpicture}

\caption {The impact of three synthesis stages. The horizontal axis represents emotion recognition accuracy and deep red arrows indicate the error at each stage. The prompt audios or speech tokens utilize (1) the human speech itself, (2) truncated human speech with half length, (3) cross-speaker within TESS dataset, and (4) out-of-domain speaker in RAVDESS dataset.}
\label{fig:tts_stages}

\end{figure}
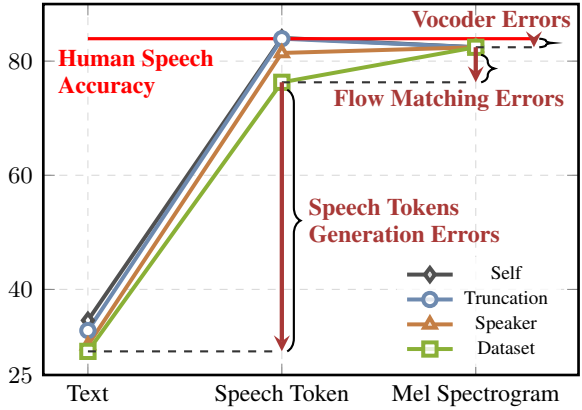

\paragraph{Synthesis gap arising from speech token error.} To simulate the transition from ideal to realistic synthesis scenarios, we employed four types of reference speech: (1) the real human speech itself, (2) truncated real speech with half length, (3) the remaining speaker in TESS dataset, and (4) out-of-domain speaker in RAVDESS dataset.
As shown in Fig. \ref{fig:tts_stages}, real speech achieves an SER accuracy of 83.93\%, while Mel-spectrogram reconstruction reaches 82.43\%, indicating that the vocoder has a negligible impact on emotion recognition. When using ground-truth tokens with the original audio as a reference, the accuracy reaches 84.07\%, slightly surpassing the SER accuracy in real human speech. Averaged across the four reference settings, flow matching introduces slightly more error than the vocoder but remains a minor factor.

In contrast, driving the synthesis with text inputs utilizing the AR language model causes a drastic performance drop to approximately 30\%. Even using the target audio itself as a reference yields only 34.57\%, improving to just $\sim$ 50\% after manually filtering for emotional expressiveness. While Mel-based reconstruction remains unaffected by reference selection, accuracy for text-based generation declines progressively from ideal to realistic reference conditions. Consequently, our results conclusively show that the discrepancy between AR-generated tokens and ground-truth tokenizer outputs is the dominant factor degrading emotion recognition performance in synthesized speech.

\section{Conclusion}
This paper highlights a clear gap between human and synthesized speech, leading to poor emotion recognition performance in discriminative Speech Emotion Recognition (SER) models on synthesized audio. This generalization issue arises from the speech token prediction stage in synthesis, which induces a representation mismatch. Additionally, generative speech LLMs rely on textual semantics to recognize emotion, often neglecting paralinguistic features. Overall, our findings suggest that existing SER models exploit non-robust shortcuts rather than capturing intrinsic emotional features.

\section*{Limitation}
While our work offers insights into the generalization gap between human and synthesized speech in emotion recognition, several limitations remain.
Firstly, to ensure the validity of our evaluation, we relied on manual annotation to filter out synthesized utterances that failed to convey the target emotion. This process introduces potential human subjectivity and restricts the scale of our synthetic testing data. 
Secondly, although we attempted to bridge the domain gap using Domain Adversarial Neural Networks, our results indicate that this approach yields limited generalization to unseen synthesis models. Future work should focus more on pretraining better representation models.

\section*{Ethical Considerations}

\paragraph{Potential Risks.}
Our work involves the analysis of synthesized speech with emotional paralinguistic features. We acknowledge that advancements in emotional TTS and S2S models carry potential risks, including the creation of deepfakes for fraud, impersonation, or manipulation. However, the primary goal of this research is to critically evaluate the limitations of current Speech Emotion Recognition (SER) models in detecting and understanding these synthetic artifacts. By highlighting the ``human-synthesis gap'' and the vulnerabilities of SER models to token-prediction errors, our work contributes to the development of more robust detection systems and safer AI interactions.

\paragraph{Human Annotation and Fair Compensation.}
To ensure the validity of our synthesized datasets, we recruited four human annotators to manually verify the emotional expressiveness of the generated audio clips.
\begin{itemize}
    \item \textbf{Recruitment and Demographics:} The annotators were recruited from graduate students from our research group. They are proficient in English with normal hearing capabilities.
    \item \textbf{Task and Instructions:} Annotators were instructed to listen to synthesized audio samples and determine if the target emotion (e.g., angry, happy, fearful) was clearly perceptible. Samples deemed ambiguous were discarded. 
    \item \textbf{Compensation:} All participants were compensated for their time. The payment rate was set at ~¥22.5 per hour, which exceeds the local minimum wage in China.
    \item \textbf{Consent and Data Privacy:} We obtained informed consent from all annotators. They were informed that the task involved listening to emotional speech (which may include negative emotions like anger or fear) and that they could withdraw from the study at any time without penalty. No personally identifiable information about the annotators was collected or stored in the final dataset.
\end{itemize}

\paragraph{Data Usage.}
The real human speech data used in this study comes from publicly available datasets. We strictly adhered to the usage licenses and terms of these datasets.

\section*{Acknowledgments}
This work was supported in part by the National Science Foundation of China (Nos. 62276056 and U24A20334), the Yunnan Fundamental Research Projects (No.202401BC070021), the Yunnan Science and Technology Major Project (No. 202502AD080014), the Fundamental Research Funds for the Central Universities (Nos. N25BSS054 and N25BSS094), and the Program of Introducing Talents of Discipline to Universities, Plan 111 (No.B16009).

We sincerely thank Professor Xie Chen from Shanghai Jiao Tong University for his constructive feedback on our three-stage synthesis analysis and for the stimulating discussions that helped refine the narrative and structure of this work.

\bibliography{custom}

\appendix

\section*{Appendix}
\definecolor{PromptBoxTitleColor}{RGB}{76,76,76}
\definecolor{PromptBoxColor}{RGB}{247,247,255}
\definecolor{block-gray}{gray}{0.85}

\newtcolorbox{mybackground}{colback=block-gray!50,grow to right by=0mm,grow to left by=0mm,boxrule=0pt,boxsep=0pt,breakable}

\section{Evaluation Details}
In this paper, we report the accuracy of speech emotion recognition models on diverse datasets, including both human and synthesized speech. 
Since the inference process of Emotion2vec is not affected by random seeds, we run inference only once and report the average accuracy of SER models over each dataset.
All experiments were conducted with 8 NVIDIA GeForce RTX 3090 GPUs.

We employ 4 manual annotators to evaluate all synthesized audio samples in data filtering process. The detailed instructions are shown below: 

\begin{itemize}
  \item \textbf{Task Overview}
  
    For each example, you will be given:
    \begin{itemize}
      \item an audio clip $S$ (about 10 seconds), and
      \item a target emotion label $L \in$ \{\texttt{surprised}, \texttt{angry}, \texttt{sad}, \texttt{disgusted}, \texttt{fearful}, \texttt{happy}, \texttt{neutral}\}.
    \end{itemize}

    Your goal is to judge whether the audio $S$ \textbf{saliently expresses} the target emotion $L$. Firstly, listen to the audio. Then decide YES/NO.
    \begin{itemize}
      \item Select \textbf{YES} if the dominant emotion is \textbf{consistent with $L$} and the expression is \textbf{clear and strong enough}.
      \item Otherwise, select \textbf{NO}.
    \end{itemize}

    \item \textbf{Output Format}

    For each example, selects \textbf{YES} or \textbf{NO}

    \item \textbf{Reference Cues for Each Emotion}
    \begin{itemize}
      \item \texttt{angry}: high energy, tense/pressed voice, strong stress, possible shouting.
      \item \texttt{disgusted}: clear aversion/contempt, ``ew''-like quality, scoffing tone.
      \item \texttt{fearful}: nervous/tense, unstable or trembling voice, rapid breathing, high/unstable pitch, panic-like urgency.
      \item \texttt{happy}: bright and lively tone, upward intonation, relaxed energy, possible laughter or smiling voice.
      \item \texttt{neutral}: steady, controlled delivery with minimal affect; no strong emotional coloration.
      \item \texttt{sad}: low energy, slower pace, downward intonation, heavy/flat tone, possible sighing quality.
      \item \texttt{surprised}: sudden pitch rise, short burst/exclamation, clear ``unexpected'' reaction with abrupt prosodic change.
    \end{itemize}
\end{itemize}

\section{Speech LLM Prompt for SER}
\label{appendix:prompt}
The prompt configuration utilized for Speech Large Language Model (SLM) inference is detailed below. To reproduce the results reported in the Fig. \ref{fig:SLM}, please omit the first two rules within the 'Rules' section. We employed the full prompt to investigate whether prompt engineering can mitigate the inherent bias in SLMs toward semantic content over emotional prosody during affective judgment.

\begin{mybackground}
You are an audio emotion classification model. \\

Your task:
Given the following audio input, classify the speaker’s emotion. \\

Emotion categories (choose exactly ONE):\\
- angry  \\
- disgust  \\
- fearful  \\
- happy  \\
- neutral  \\
- sad  \\
- surprised  \\

Rules:\\
1. Completely ignore the textual content of the speech. The transcript may be misleading or emotion-neutral.\\
2. Judge emotion only from acoustic characteristics, not semantics.\\
3. Do NOT default to "neutral" when uncertain. Instead, choose the closest non-neutral category unless the audio is clearly flat, monotone, and emotionless.\\
4. Output only the emotion label (all lowercase, no punctuation).\\
5. Do not output explanations unless explicitly requested.
\\

Output format:\\
<emotion\_label>
\end{mybackground}

\section{Speech Synthesis in S2S LLMs}
\label{appendix:s2s}
We employ Kimi-Audio and GLM-4-Voice to synthesize speech with salient emotional features for testing the classification ability on S2S LLMs synthetic data. Unlike TTS models which explicitly control emotion through prompt speech and natural language instructions, S2S LLMs generate speech in a conversational manner. During the interaction process, however, we find that the emotional expression in the first-turn response is insufficient. Hence, we introduce additional prompts in subsequent turns, such as `Repeat the previous answer and speak with a clearly angry tone'. Through this iterative prompting strategy, we are able to obtain audio samples with more pronounced emotional features in the second or third-turn response. These audio samples are manually filtered and subsequently used as synthetic data for testing the emotion classification models.

\begin{table*}[t]
    \centering
    \resizebox{\textwidth}{!}{
    \begin{tabular}{lcccccccccc}
        \toprule
        & \multicolumn{5}{c}{\textbf{Human Speech}} & \multicolumn{4}{c}{\textbf{Synthesized Speech}} \\
        \cmidrule(lr){2-6} \cmidrule(lr){7-10}
        \multicolumn{1}{c}{\textbf{Method}} & \multicolumn{5}{c}{\textbf{In Domain}} & \multicolumn{2}{c}{\textbf{In Domain}} & \multicolumn{2}{c}{\textbf{OOD}} \\
        \cmidrule(lr){2-6} \cmidrule(lr){7-8} \cmidrule(lr){9-10}
        & IEMOCAP & RAVDESS & TESS & MELD & CREMA-D & IEMOCAP & TESS & CREMA-D $\dagger$ & CREMA-D $\ddagger$ \\
        \midrule
        Emotion2vec & 71.13 & 66.35 & 99.64 & 50.15 & 78.43 & 89.20 & 88.27 & 32.77 & 49.38 \\
        Qwen2.5-omni-3B & 84.24 & 79.81 & 99.64 & 56.44 & 82.24 & 96.12 & 96.43 & 68.92 & 57.07 \\ 
        \bottomrule
    \end{tabular}
    }
    \caption{Qwen2.5-omni-3B results training with mixed datasets.}
    \label{tab:qwen-omni}
\end{table*}

\begin{figure}[t]
  \includegraphics[width=0.48\linewidth]{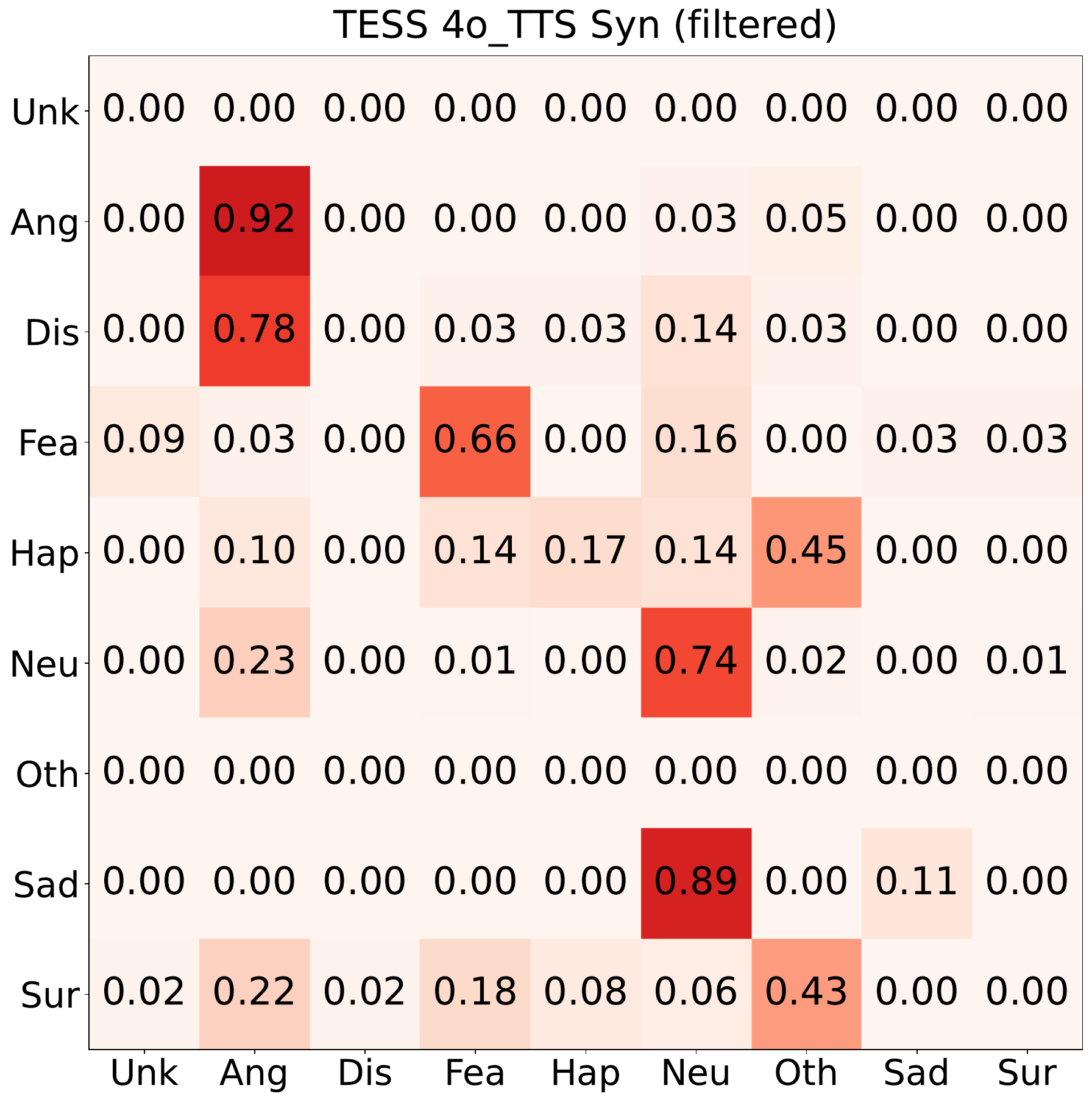} \hfill
  \includegraphics[width=0.48\linewidth]{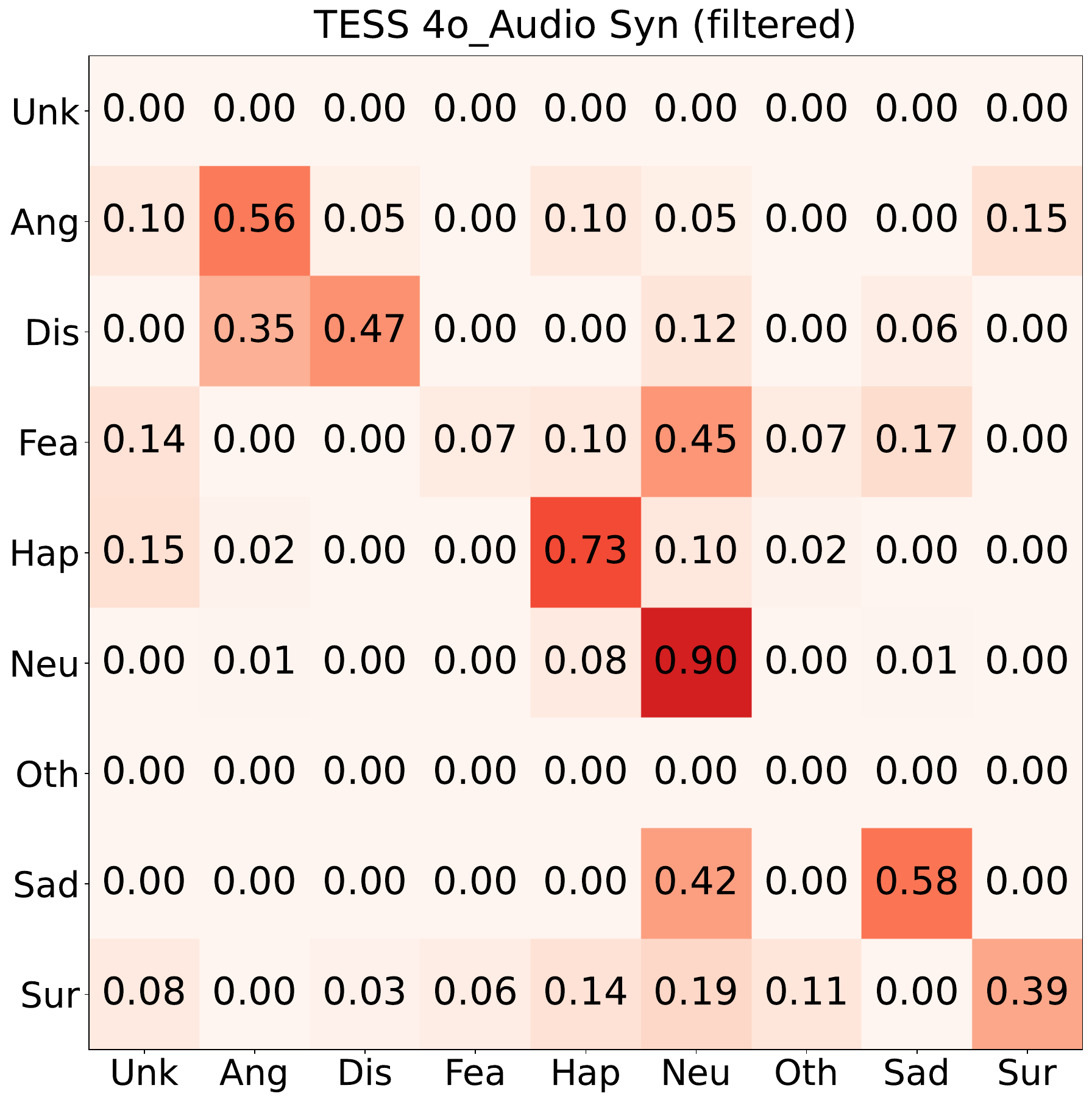} 

  \caption {Confusion matrix of speech emotion recognition results. All samples are synthesized by GPT-4o TTS and GPT-4o Audio.}
  \label{fig:gpt-results}
\end{figure}

\section{Emotion2vec Performance on Commercial Synthesized Models}
\label{appendix:commercal}
As shown in Fig. \ref{fig:gpt-results}, we present a detailed confusion matrix for speech emotion recognition on the TESS dataset synthesized by two commercial systems: GPT-4o-TTS and GPT-4o-Audio. Despite the strong synthesis ability of commercial models and the salient emotional features of filtered audio samples, the classification performance of Emotion2vec remains unreliable.

\section{Analysis of Three TTS Stages}

\definecolor{acadBlue}{RGB}{110, 138, 175}
\definecolor{acadRed}{RGB}{168, 56, 54}
\definecolor{acadGreen}{RGB}{138, 176, 54}
\definecolor{acadBrown}{RGB}{196, 126, 68}
\definecolor{darkGrey}{RGB}{80, 80, 80}

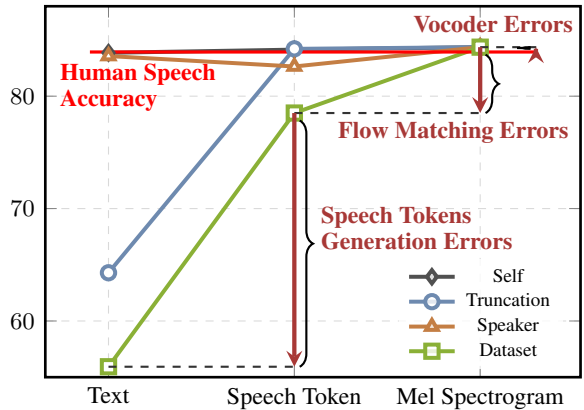
\begin{figure}[t]
\centering
\begin{tikzpicture}
    \tikzset{
        errorbrace/.style={
            decorate,
            decoration={brace, amplitude=5pt, raise=2pt}, 
            thick, 
            black 
        }
    }

    \begin{axis}[
    width=1.12\linewidth, 
    height=6.5cm,
    grid=major,
    grid style={dashed, gray!30},
    xlabel style={font=\small, yshift=-3pt},
    ylabel style={font=\small},
    xtick={1,2,3},
    xticklabels={Text, Speech Token, Mel Spectrogram},
    xticklabel style={font=\footnotesize},
    tick label style={font=\footnotesize},
    ymin=55, ymax=88,
    ytick={60,70,80},
    axis lines=box,
    axis line style={line width=1pt},          
    every axis plot/.append style={             
        line width=1.5pt,
        mark size=2.5pt
    },
    legend style={
        at={(0.66,0.32)},
        anchor=north west,
        draw=none,
        fill=white,
        fill opacity=0.8,
        text opacity=1,
        font=\small,
        legend columns=1,
        column sep=2pt,
        nodes={scale=0.8, transform shape}
    },
    ]

    \addplot[color=darkGrey, mark=diamond*, mark options={fill=white}] coordinates {
        (1, 83.86) (2, 84.14) (3, 84.36)
    };
    \addlegendentry{Self}

    \addplot[color=acadBlue, mark=*, mark options={fill=white}] coordinates {
        (1, 64.29) (2, 84.21) (3, 84.36)
    };
    \addlegendentry{Truncation}

    \addplot[color=acadBrown, mark=triangle*, mark options={fill=white}] coordinates {
        (1, 83.57) (2, 82.64) (3, 84.36)
    };
    \addlegendentry{Speaker}

    \addplot[color=acadGreen, mark=square*, mark options={fill=white}] coordinates {
        (1, 55.93) (2, 78.50) (3, 84.36)
    };
    \addlegendentry{Dataset}

    \addplot[very thick, red, domain=0.9:3.3, samples=2] {83.93} 
        node[below, at start, align=left, xshift=18pt, font=\bfseries\small, red] {Human Speech\\Accuracy};

    \coordinate (D_Token) at (axis cs:2, 78.50);   
    \coordinate (B_Token) at (axis cs:1, 55.93); 
    \coordinate (M_Token) at (axis cs:2, 55.93); 
    
    \coordinate (D_Mel)   at (axis cs:3, 84.36);
    \coordinate (B_Mel)   at (axis cs:2, 78.50);
    \coordinate (M_Mel) at (axis cs:3, 78.50); 

    \coordinate (D_Human) at (axis cs:3.3, 84.21);
    \coordinate (B_Human) at (axis cs:3, 84.36);
    \coordinate (M_Human) at (axis cs:3.3, 84.36);

    \end{axis}

    \draw[thick, dashed, black!80] (B_Token) -- ++(2.5, 0);
    \draw[errorbrace] ($(D_Token)+(0,-0.1)$) -- (M_Token) 
        node[midway, right=6pt, align=left, yshift=6pt, font=\bfseries\small, color=acadRed] 
        {Speech Tokens \\ Generation Errors};

    \draw[thick, dashed, black!80] (B_Mel) -- ++(2.5, 0);
    \draw[errorbrace] ($(D_Mel)+(0,-0.1)$) -- (M_Mel) 
        node[midway, below=10pt, xshift=-10pt, align=left, font=\bfseries\small, color=acadRed] 
        {Flow Matching Errors};

    \draw[thick, dashed, black!80] (B_Human) -- ++(0.8, 0);
    \draw[errorbrace] (D_Human) -- (M_Human)
        node[midway, above=2pt, xshift=-16pt, align=left, font=\bfseries\small, color=acadRed] 
        {Vocoder Errors};
        
    \draw[->, >=stealth, ultra thick, acadRed] (D_Token) -- (M_Token);
    \draw[->, >=stealth, ultra thick, acadRed] (D_Mel) -- (M_Mel);
    \draw[->, >=stealth, ultra thick, acadRed] (D_Human) -- (M_Human);

\end{tikzpicture}

\caption {The impact of three synthesis stages of IndexTTS2.}
\label{fig:appendix_tts_stages}

\end{figure}

As shown in Fig. \ref{fig:appendix_tts_stages}, we present additional analysis on the impact of three synthesis stages of IndexTTS2 \citep{zhou2025indextts2}. Experimental results also indicate that \textit{speech token generation errors} domain the synthesis gap.

\section{Training LALMs for SER}
\label{appendix:lalms}
Furthermore, we have included new results for Qwen2.5-Omni-3B \citep{Xu2025Qwen25OmniTR} trained with mixed datasets. As shown in Table \ref{tab:qwen-omni}, fine-tuning Qwen2.5-Omni significantly outperforms the emotion2vec model and successfully mitigates the issue where S2S LLMs rely on textual features rather than acoustic emotional features for recognition. However, the model still struggles with generalization: \textit{even with Qwen2.5-Omni, performance remains poor on OOD scenarios} (unseen TTS models or datasets).

\section{Emotion Recognition Relevance}
\definecolor{acadBlue}{RGB}{110, 138, 175}
\definecolor{acadBrown}{RGB}{196, 126, 68}

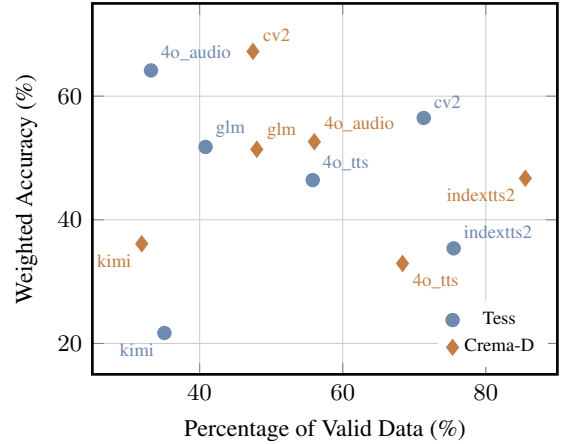
\begin{figure}[t]
\centering
\begin{tikzpicture}
\begin{axis}[
    width=1\linewidth,
    height=6.5cm,
    xlabel={Percentage of Valid Data (\%)},
    ylabel={Weighted Accuracy (\%)},
    xmin=25, xmax=90,
    ymin=15, ymax=75,
    grid=both,
    axis lines=box,
    axis line style={line width=1pt},
    major grid style={line width=0.3pt, draw=gray!40},
    minor grid style={line width=0.2pt, draw=gray!20},
    tick label style={font=\small},
    label style={font=\small},
    legend style={
        at={(0.98,0.02)},
        anchor=south east,
        font=\scriptsize,
        draw=none,
    }
]

\addplot[
    only marks,
    mark=*,
    mark size=2.5pt,
    acadBlue
] coordinates {
    (71.32,56.48)  
    (75.50,35.38)  
    (35.07,21.69)  
    (40.82,51.79)  
    (55.80,46.42)  
    (33.20,64.16)  
};
\addlegendentry{Tess}

\node[acadBlue, font=\scriptsize, anchor=south west] at (axis cs:71.32,56.48) {cv2};
\node[acadBlue, font=\scriptsize, anchor=south west] at (axis cs:75.50,35.38) {indextts2};
\node[acadBlue, font=\scriptsize, anchor=north east] at (axis cs:35.07,21.69) {kimi};
\node[acadBlue, font=\scriptsize, anchor=south west] at (axis cs:40.82,51.79) {glm};
\node[acadBlue, font=\scriptsize, anchor=south west] at (axis cs:55.80,46.42) {4o\_tts};
\node[acadBlue, font=\scriptsize, anchor=south west] at (axis cs:33.20,64.16) {4o\_audio};

\addplot[
    only marks,
    mark=diamond*,
    mark size=3pt,
    acadBrown
] coordinates {
    (47.45,67.23)  
    (85.52,46.71)  
    (31.90,36.13)  
    (47.99,51.40)  
    (68.36,32.94)  
    (56.03,52.63)  
};
\addlegendentry{Crema-D}

\node[acadBrown, font=\scriptsize, anchor=south west] at (axis cs:47.45,67.23) {cv2};
\node[acadBrown, font=\scriptsize, anchor=north east] at (axis cs:85.52,46.71) {indextts2};
\node[acadBrown, font=\scriptsize, anchor=north east] at (axis cs:31.90,36.13) {kimi};
\node[acadBrown, font=\scriptsize, anchor=south west] at (axis cs:47.99,51.40) {glm};
\node[acadBrown, font=\scriptsize, anchor=north west] at (axis cs:68.36,32.94) {4o\_tts};
\node[acadBrown, font=\scriptsize, anchor=south west] at (axis cs:56.03,52.63) {4o\_audio};

\end{axis}
\end{tikzpicture}

\caption {The horizontal axis represents the proportion of data where humans perceive significant emotional expression, while the vertical axis indicates the Emotion2vec SER accuracy.}
\label{fig:scatter_diagram}

\end{figure}

We visualized the relationship between the importance of emotion expression in synthetic models and the accuracy of emotion recognition of Emotion2vec. 
As shown in Fig. \ref{fig:scatter_diagram}, the x-axis represents the proportion of synthetic audio in which human annotators believe the emotion is significantly expressed. The y-axis represents the emotion recognition accuracy of Emotion2vec. The visualization in Fig. \ref{fig:scatter_diagram} shows that there is no strong positive correlation between the two metrics.

\section{The Significance of Synthetic Data in Ideal Scenarios}

\begin{figure}[tb]
\centering
\resizebox{\linewidth}{!}{
    \begin{tikzpicture}
      
      
      \draw[fill=mygrey, opacity=0.8, draw=black, thick]
        (-5, -0.5) 
        .. controls (-6, -4.5) and (6.2, -5.0) .. (5, -0.5)
        .. controls (4, 3.5) and (-4, 3.5) .. (-5, -0.5)
        -- cycle; 
      \node at (0.5, 1.7) {\Huge $\mathcal{P}^*$};

      \draw[fill=myblue, opacity=0.8, draw=blue!40!white, thick] (-1.6,-0.6) ellipse (3.2cm and 1.8cm);
      \node at (-2.0, -0.1) {\Large Human Speech};
      \node at (-2.0, -0.8) {\Huge $\mathcal{P}$};

      \draw[fill=myorange, opacity=0.8, draw=orange!80!black, thick, rotate around={20:(2.5,-0.8)}] (3.0,-0.4) ellipse (3.2cm and 1.6cm);
      \node at (5.0, 0.8) {\Huge $\mathcal{P}'$};

      \draw[fill=mypurple, opacity=0.6, draw=purple!60!black, thick, rotate around={20:(2.5,-0.8)}] (2.5,-0.4) ellipse (2.2cm and 1.3cm);
      \node at (2.6, -0.4) {\Huge $\mathcal{P}'_T$};
      
      \node at (1.1, -2.8) {\Large $\mathcal{P}_T' \approx$ long-tail($\mathcal{P}^*$)};
      
    \end{tikzpicture}
}
\caption{The significance of synthetic data.}
\label{fig:distribution_venn}
\end{figure}
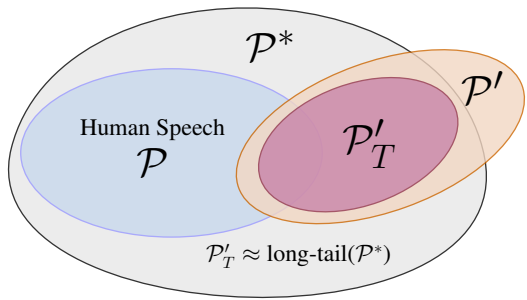

We investigate the unique value of synthetic data in speech understanding tasks. For instance, can synthetic audio, by supplementing long-tail distribution data, enable neural networks to uncover more intrinsic features relevant to the task?

As shown in  Fig. \ref{fig:distribution_venn}, consider a scenario where authentic speech-emotion data follows a common distribution $\mathcal{P}$. Conversely, certain synthetic audio-emotion data, influenced by the generative process, may form a less common distribution $\mathcal{P}'$. Within this, a subset $\mathcal{P}_T'$ represents synthetic data where emotions remain discernible to humans. We posit that $\mathcal{P}_T'$ effectively constitutes the long-tail portion of the true, complete distribution $\mathcal{P}^*$. Consequently, we investigate whether leveraging the more comprehensive distribution formed by combining $\mathcal{P}$ and $\mathcal{P}_T'$ allows for the discovery of more intrinsic features for emotion recognition, thereby transcending the limitations of relying solely on the common distribution $\mathcal{P}$.
Furthermore, we explore the potential of leveraging scaling laws to enhance the model's representation of the true distribution $\mathcal{P}^*$ through the use of massive pseudo-labeled synthetic data. 
Recent work has demonstrated that in language modeling tasks \citep{zeng2024glm}, utilizing large-scale synthetic data facilitates the learning of more robust speech-text alignment relationships. 
However, the answers to these questions are still unclear in the domain of speech understanding.

\end{document}